\pdfoutput = 1

\documentclass[12pt]{article}
\usepackage{amsmath,amssymb}
\usepackage{times}
\usepackage{graphicx}
\usepackage{color}
\usepackage{multirow}
\usepackage[authoryear]{natbib}
\usepackage{rotating}
\usepackage{bbm}
\usepackage{latexsym}
\usepackage{dcolumn}
\usepackage{bm}
\usepackage{xcolor}

\newcommand{\tr}{^\intercal}
\newcommand{\bb}{\bm{b}}
\newcommand{\bv}{\bm{v}}
\newcommand{\bz}{\bm{z}}
\newcommand{\bW}{\bm{W}}
\newcommand{\bq}{\bm{q}}
\newcommand{\bJ}{\bm{J}}
\newcommand{\bU}{\bm{U}}
\newcommand{\bB}{\bm{B}}
\newcommand{\bc}{\bm{c}}
\newcommand{\bLambda}{\bm{\Lambda}}
\DeclareMathOperator{\erf}{erf}

\DeclareMathOperator{\floor}{floor}

\newcommand{\captionfonts}{\normalsize}

\makeatletter  
\long\def\@makecaption#1#2{%
  \vskip\abovecaptionskip
  \sbox\@tempboxa{{\captionfonts #1: #2}}%
  \ifdim \wd\@tempboxa >\hsize
    {\captionfonts #1: #2\par}
  \else
    \hbox to\hsize{\hfil\box\@tempboxa\hfil}%
  \fi
  \vskip\belowcaptionskip}
\makeatother   

\begin{document}
\hspace{13.9cm}

\ \vspace{2mm}\\

{\noindent\LARGE Restricted Boltzmann Machines as Models of \\ Interacting Variables}

\ \\
{\bf \large Nicola Bulso$^{1,2}$, Yasser Roudi$^{1}$}\\
{$^{\displaystyle 1}$Kavli Institute for Systems Neuroscience and Centre for Neural Computation, Norwegian University of Science and Technology, Olav Kyrres Gate 9, 7491 Trondheim, Norway}\\
{$^{\displaystyle 2}$ SISSA—Cognitive Neuroscience, Via Bonomea 265, 34136 Trieste, Italy}\\
%


\thispagestyle{empty}
\markboth{}{NC instructions}
\ \vspace{-0mm}\\
%
\begin{center} {\bf Abstract} \end{center}
We study the type of distributions that Restricted Boltzmann Machines (RBMs) with different activation functions can express by investigating the effect of the activation function of the hidden nodes on the marginal distribution they impose on observed binary nodes. We report an exact expression for these marginals in the form of a model of interacting binary variables with the explicit form of the interactions depending on the hidden node activation function. We study the properties of these interactions in detail and evaluate how the accuracy with which the RBM approximates distributions over binary variables depends on the hidden node activation function and on the number of hidden nodes. When the inferred RBM parameters are weak, an intuitive pattern is found for the expression of the interaction terms which reduces substantially the differences across activation functions. We show that the weak parameter approximation is a good approximation for different RBMs trained on the MNIST dataset. Interestingly, in these cases, the mapping reveals that the inferred models are essentially low order interaction models.

\section{Introduction}\label{introduction}
Many of the principal phenomena discovered and studied in the theory of neural computation, artificial or biological, e.g. the formation of attractors using Hebbian learning, the ability of approximating any function using feedforward networks or modelling probability distribution using Boltzmann Machines, have been initially discovered by studying networks of binary units \citep{hertz1991introduction}. Although the neural architectures involved have been shown to exhibit these phenomena when other activation functions instead of binary units are used, it is only in recent years that attention has been given to the activation function as a crucial factor that can determine the performance of the network both quantitatively and qualitatively. Notable examples are the outperformance of Rectified Linear Units (ReLU)/Threshold-Linear (TL) compared to e.g. binary units in deep neural networks trained with back-propagation \citep{nair2010rectified}, the general dependence of the performance of gradient based learning methods to the smoothness of the activation function \cite{panigrahi2019effect}, and the significantly smaller gap between optimal capacity and what is achieved via Hebbian learning \cite{schonsberg2021efficiency} in networks with ReLU/TL compared to binary networks. In this paper, we focus on understanding the role of the hidden layer activation function on the distributions that can be represented by the Restricted Boltzmann Machine (RBM).

The RBM, originally introduced by the name of Harmonium \citep{smolensky1986information}, is a bipartite undirected graphical model where only nodes belonging to one class are visible. The interactions are allowed only between nodes belonging to different classes, but not within the same class. In fact, the RBM owes its name to the fact that it represents a simplified version of the Boltzmann Machine (BM) \citep{ackley1985learning}. As compared to the BM, the RBM is simpler, but yet very powerful \citep{le2008representational}, and it can be trained with relative ease and efficiency \citep{hinton2014boltzmann}. Given its attractive features, it is widely employed in applications as a generative model of data distributions \citep{salakhutdinov2007restricted}, but also in classification tasks \citep{larochelle2012learning}. RBMs have also been used as building block of deep neural architectures, fostering the rise of deep learning \citep{hinton2006fast,roudi2015learning}. In neuroscience, RBMs have been used for modelling correlations between neuronal firings, a problem of particular importance in neural data analysis \citep{averbeck2006neural,latham2013population}, in particular for modelling correlations beyond pairwise \citep{koster2014modeling}. They have also been used for classifying stimuli impinging on the retina based on the evoked neural activity \citep{zanotto2017modeling,gardella2018blindfold} and for automatically categorising and discovering novel sleep stages in mice \citep{katsageorgiou2018novel}, among other examples.

Stochastic units in the standard RBM are binary and follow a Bernoulli conditional distribution \citep{hinton2012practical,fischer2014training} and previous works have shown that this standard RBM is a universal approximator of discrete distributions \citep{le2008representational, martens2013representational}. Many variants to the standard binary RBM have been developed, including RBMs with Gaussian units \citep{freund1992unsupervised}, Poisson units \citep{gehler2006rate}, units following distributions belonging to the exponential family \citep{welling2005exponential,ravanbakhsh2016stochastic} and Rectified Linear Unit (ReLU) \citep{nair2010rectified}. Despite this, the direct relationship between different aspects of the architecture, e.g. the properties of the hidden nodes and the connectivity matrix, and the distribution that the network expresses is often hard to understand. This is indeed a problem shared with most deep neural architectures \citep{caruana2015intelligible, foerster2017input, lipton2018mythos}, and addressing this problem for the RBMs can serve a crucial step towards addressing it in the case of deep architectures.

In this paper we consider RBMs with binary visible units and investigate how their representational power depends on the architectural characteristics of the machines by studying how the marginal distribution expressed by different RBMs map into distributions of interacting binary variables over the visible nodes, also called {\itshape spin models} in the statistical physics community. Specifically, in section 2 and 3, we derive an exact mathematical expression for this mapping and describe how it can provide insights into the way an RBM works and what kind of distributions it can represent as one varies the activation function and the number of hidden nodes. In section 4, we test the mapping with numerical simulations for a number of simple models and apply it to the study of the MNIST dataset. We compare the performances of these different RBMs and discuss how the binary interactive models they express over the visible nodes relate to each other. We conclude in section 5 by summarising the results and stressing the significance and potential of our findings.


\section{Mapping RBMs into models of interacting binary variables}
We consider a stochastic neural network with two layers: a visible layer containing $N$ nodes whose state is denoted by $\bv=(v_1,\cdots, v_N)$, and a hidden layer containing $M$ nodes, whose state is denoted by  $\bz=(z_1,\cdots,z_M) $. In the following, we label elements of the visible layer with roman subscripts, e.g. $v_i$, and elements of the hidden layer with greek subscripts, e.g. $z_\mu$. The visible and the hidden nodes interact through undirected pairwise connections represented by the $N$-by-$M$ weight matrix $\bW$. The joint probability distribution of such a generalised RBM can be written as 
\begin{equation}
P(\bv,\bz) = \frac{1}{Z} \exp\left[\bb\tr \bv + \bv\tr \bW\bz -U(\bz)\right],
\label{modeldist}
\end{equation}
where $Z$ is the normalisation or partition function, $\bb=(b_1,\cdots,b_N)$ are bias terms acting on the visible nodes, while the hidden nodes are subjected to the potential $U(\bz)$. Visible nodes are binary variables, $v_{i} \in \{0,1\}$, and they are not directly interacting with each other. Hidden nodes, instead, can be discrete or continuous variables and interacting with each other or not, all depending on the form the chosen potential $U(\bz)$. 

The model described above and in Eq.\ \eqref{modeldist} is a generalisation of the standard Restricted Boltzmann Machine (RBM): it retrieves the standard RBM when the potential $U(\bz)$ is chosen such that hidden nodes are binary, their activation function follows a step function and no hidden-hidden connections are permitted.  By exploiting Bayes rule, it is easy to see that visible nodes are conditionally independent given the values of the hidden nodes, namely that the conditional probability $P(\bv|\bz)$ factorises over the visible nodes, i.e. $P(\bv|\bz) = \prod_i P(v_i |\bz)$. On the other hand, the hidden nodes are conditionally independent given the visibles only if the potential can be decomposed into a sum of $M$ independent terms, namely $U(\bz) = \sum_{\mu} U_{\mu}(z_\mu)$. We refer to this particular choice of potential as a {\itshape separable potential}. 

The marginal distribution of the observed nodes, $P(\bv)$, for the generalised RBM described in Eq.\ \eqref{modeldist} can be written as
\begin{equation}
P(\bv) = \frac{1}{Z}\exp\left[\bb\tr \bv + \log\int d\bz \exp (\bv\tr \bW \bz -U(\bz))\right].
\label{eqn:RBMdistribution}
\end{equation}
In order to understand the effect of the choice of a particular architecture on the representational power of the corresponding RBM, namely the subset of distributions that the RBM is able to capture, we describe in the following section how the marginal distribution $P(\bv)$ changes by varying the potential $U$. We do this by expressing the marginal distribution in Eq.\ \eqref{eqn:RBMdistribution} into a model of interacting binary variables,
\begin{equation}\label{eqn:exponential}
P(\bv) = \frac{1}{Z'}\exp\left( \sum_{k_1} I^{(1)}_{k_1}v_{k_1} + \sum_{k_{1} < k_{2}} I^{(2)}_{k_1,k_2}v_{k_1}v_{k_2} + \ldots  + I^{(N)}_{1,2,\ldots,N} \prod_{k=1}^N v_{k} \right),
\end{equation}
where $I^{(s)}_{k_{1},k_{2},\ldots,k_{s}}$ represents the $s$th order interaction term between visible nodes $k_{1},k_{2},\ldots,k_{s}$ of the original RBM.

We derive the equivalence between RBMs and models of interacting variables by expressing the argument of the exponential in Eq.\ \eqref{eqn:RBMdistribution} as a sum of interaction terms of any order, as Eq.\ \eqref{eqn:exponential}. Finding the parametrisation that the RBM employs to represent the interaction parameters in Eq.\ \eqref{eqn:exponential} allows us to understand how the RBM exploits its limited resources to approximate a generic probability distribution over binary variables.


\subsection{The special case of the quadratic potential}\label{sec:maci}

Eq.\ \eqref{eqn:RBMdistribution} can be rewritten in a more convenient form as
\begin{equation}\label{eqn:convenient}
P(\bv) = \frac{1}{Z'}\exp\left[\bb\tr \bv + \log\int d\bz \exp (\bv\tr \bW \bz)\,\rho (\bz)\right],
\end{equation}
where $Z' = Z/\int d\bz \exp[-U(\bz)]$ and
\begin{equation}\label{eqn:density}
\rho (\bz) = \frac{\exp[-U(\bz)]}{\int d\bz \exp[-U(\bz)]}.
\end{equation}
In what follows, we assume that the potential $U(\bz)$ satisfies the requirements for $\rho(\bz)$ to be a probability density. Denoting the cumulant generating function of the distribution $\rho$ by $K$, that is
\begin{equation}
K(\bq) \equiv \log \int d\bz \exp (\bq\tr \bz)\,\rho (\bz),
\label{eq:Kdef}
\end{equation}
we can write Eq.\ \eqref{eqn:convenient} as
\begin{equation}\label{eqn:visible_moment_gen_fun}
P(\bv) = \frac{1}{Z'}\exp\left[\bb\tr \bv + K(\bW\tr \bv)\right].
\end{equation}

It has been shown in \citet{marcinkiewicz1939propriete} that the cumulant generating function cannot be a finite-order polynomial of degree greater than two (see also \citet{rajagopal1974some} for generalizations). This result, combined with the idempotence property of the parametrisation, namely $v_i^n = v_i$ for any positive integer $n$, implies that the argument of the exponential in Eq.\ \eqref{eqn:visible_moment_gen_fun} will either be a function of only single and pairwise terms, namely $v_i$, $\forall i$, and $v_i v_j$, $\forall i<j$, respectively, or alternatively it will involve all possible $2^N -1$ interactions. As a result, any RBM, as defined in the previous section, can either represent pure pairwise models where variables are interacting only through pairwise interactions or it will describe a model with all orders of interactions\footnote{It is worth noting that in the case of the generalised Hopfield model studied in \cite{barra2017phase}, a similar result can be derived by applying Marcinkiewicz's theorem \citep{marcinkiewicz1939propriete} to the cumulant generation functional representation of the $u$ function defined in Eq.\ (2) of that paper.}. This implies, for instance, that there is no such choice of potential that leads to an RBM describing a system where variables are interacting only through three-body terms, {\em for every choice of $\bW$}. For particular choices of $\bW$, however, it is still possible to limit the largest order of interaction terms for an RBM with that $\bW$. The special case of $U(\bz)$ leading to a model with only pairwise interactions is a quadratic potential, which corresponds to $\rho(\bz)$ following a Gaussian distribution.


\subsection{An exact mapping between RBMs and models of interacting binary variables}
In this section, we are going to derive the mapping between Eqs.\ \eqref{eqn:RBMdistribution} and \eqref{eqn:exponential} by finding the parametrisation employed by the RBM for describing the interaction terms. Since the visible variables are binary and $v_i \in \{0,1\}$ $\forall i$, it is possible to express the cumulant generating function $K(\bW\tr \bv)$ in Eq.\ \eqref{eqn:visible_moment_gen_fun} as a sum of mutually exclusive terms, that is
\begin{align}\label{eqn:cum_expansion}
\begin{split}
& K (\bW\tr \bv) =  K(0) \prod_{k} (1-v_k) + \sum_{k_1} K\left(\bW\tr_{k_1}\right) v_{k_1} \prod_{k\neq k_1} (1-v_k)  + \\
& + \sum_{k_1<k_2} K\left(\bW\tr_{k_1} + \bW\tr_{k_2}\right) v_{k_1}\!v_{k_2}  \prod_{k \neq k_1,k_2} (1-v_k) + \ldots  \\
& = \sum_{l =0}^N \sum_{k_1<\cdots < k_l} \!\!\!\! K \left(\sum_{j = 1}^{l} \bW\tr_{k_j}\right) v_{k_1}\cdots v_{k_l}  \!\!\!\! \prod_{k\neq (k_1,...,k_l)} \!\!\!\! (1-v_k),
\end{split}
\end{align}
where $\bW\tr_{k}$ is the $k$-th column of the transposed weight matrix $\bW\tr$. Moreover, the term $\prod_{k\neq (k_1,...,k_l)}(1-v_k)$ can be further expanded as
\begin{equation}\label{eqn:prod_expansion}
\prod_{k\neq (k_1,...,k_l)} \!\!\!\! (1-v_k) = \sum_{p = 0}^{N-l} (-1)^{p} \!\!\!\! \sum_{\substack{k_{l+1}<\cdots<k_{l+p} \\
\neq (k_1,...,k_l)
}} \!\!\!\! v_{k_{l+1}}\cdots v_{k_{l+p}}.
\end{equation}
Substituting Eq.\ \eqref{eqn:prod_expansion} into Eq.\ \eqref{eqn:cum_expansion}, it follows that
\begin{equation}\label{eqn:K1_expansion}
K (\bW\tr \bv) = \sum_{l =0}^N \sum_{p = 0}^{N-l} (-1)^{p} \!\!\!\! \sum_{\substack{k_1<\cdots<k_l ; \\ k_{l+1}<\cdots<k_{l+p} \\ \neq (k_1,...,k_l)}} \!\!\!\! K \left(\sum_{j = 1}^{l} \bW\tr_{k_j}\right) v_{k_1}\cdots v_{k_{l+p}}.
\end{equation}
In the last equation, it is possible to rearrange the order of the summation to make each addend symmetric under permutations of all indices $(k_1,...,k_{l+p})$: this is achieved by summing over all possible $l+p$ indices out of $N$ and then over all possible $l$ indices out of the selected $l+p$ indices, instead of summing over all possible $l$ indices out of $N$ followed by a sum over all possible $p$ indices out of the remaining $N-l$. This means exploiting the following equivalence
\begin{equation}
\!\!\!\! \sum_{\substack{k_1<\cdots<k_l ; \\ k_{l+1}<\cdots<k_{l+p} \\ \neq (k_1,...,k_l)}} \!\!\!\! K \left (\sum_{j = 1}^{l} \bW\tr_{k_j}\right) v_{k_1}\cdots v_{k_{l+p}} = \sum_{k_1<\cdots<k_{l+p}} \sum_{\substack{j_1<\cdots<j_{l} \\ =1}}^{l+p} K\left(\sum_{i = 1}^{l} \bW_{k_{j_i}}\tr\right) v_{k_1}\cdots v_{k_{l+p}} .
\end{equation}
Finally, by replacing the latter expression into Eq.\ \eqref{eqn:K1_expansion} and by substituting the sum over $l$ with a sum over $s = l+p$, we obtain
\begin{equation}\label{eqn:K_expansion}
 K (\bW\tr \bv) = \sum_{s =1}^N \sum_{k_1<\cdots<k_s} I^{(s)}_{k_1,k_2,...,k_s} v_{k_{1}}\cdots v_{k_{s}} ,
\end{equation}
where 
\begin{equation}\label{eqn:general_int_terms}
I^{(s)}_{k_1,k_2,...,k_s}  = \sum_{p = 0}^{s-1} (-1)^{p} \!\!\!\! \sum_{j_1<\cdots<j_{s-p}=1}^s \!\!\!\! K\left( \sum_{i = 1}^{s-p} \bW_{k_{j_i}}\tr\right).
\end{equation}
Notice that in the last equations the term $s=0$ and $p=s$ were removed because they do not contribute to the sum, given that $K(0)=0$. 
Eq.\ \eqref{eqn:K_expansion} represents the expansion of the cumulant generating function in terms of binary variables. It is worth noticing that this expansion is a general result as it can be applied to any function whose argument is a linear combination of binary factors taking values $0$ or $1$. Given the form of the marginal distribution of the observed nodes in Eq.\ \eqref{eqn:visible_moment_gen_fun} and the expansion in Eq.\ \eqref{eqn:K_expansion}, each parameter $I^{(s)}_{k_1,k_2,...,k_s}$ in Eq.\ \eqref{eqn:general_int_terms} represents the $s$th order interaction term in the equivalent interaction model in Eq.\ \eqref{eqn:exponential}, with the only exception of the $s=1$ case where the bias term must be included.


\section{Separable potentials}
\label{sec:seppotential}
Common choices of potentials in practice are separable potentials, namely $U(\bz) = \sum_{\mu} U_{\mu}(z_{\mu})$. In this case, the distribution $\rho(\bz)$ in Eq.\ \eqref{eqn:density} factorises over the hidden nodes and, consequently, the cumulant generating function breaks down into a sum of terms. As a result, each hidden node contributes independently to every interaction term through additive factors
\begin{equation}\label{eqn:general_int_terms_factorized}
I^{(s)}_{k_1,k_2,...,k_s}  = \sum_{\mu = 1}^{M}\sum_{p=0}^{s-1} (-1)^{p} \!\!\!\! \sum_{j_1<j_2<\cdots<j_{s-p}=1}^s \!\!\!\! K\left(\sum_{l = 1}^{s-p} w_{k_{j_l},\mu}\right) .
\end{equation}
Although the compact mathematical notation might make the above expression appear a bit obscure, writing down few examples will reveal the simplicity and regularity of the scheme. Accordingly, we report below the interactions up to the four-body term, including the bias $\bb$ and using the same notation as in Eq.\ \eqref{eqn:exponential} to make the comparison straightforward
\begin{subequations}\label{eqn:4int_terms_separable}
\begin{align}
&I^{(1)}_{k_1} = b_{k_1} + \sum_{\mu} K(w_{k_1,\mu}) \label{I1k1}\\
&I^{(2)}_{k_1,k_2} = \sum_\mu \left[ K\left(w_{k_1,\mu}+w_{k_2,\mu}\right) - K(w_{k_1,\mu}) - K(w_{k_2,\mu}) \right] \\
&I^{(3)}_{k_1,k_2,k_3} = \sum_\mu \left[ K\left(\sum_j w_{k_j,\mu}\right) -\sum_{j_1<j_2} K\left(w_{k_{j_1},\mu}+w_{k_{j_2},\mu}\right)  + \sum_j K(w_{k_j,\mu})\right] \\
&I^{(4)}_{k_1,k_2,k_3,k_4} = \sum_\mu \left[ K\left(\sum_j w_{k_j,\mu}\right)  -\sum_{j_1<j_2<j_3} K(w_{k_{j_1},\mu}+w_{k_{j_2},\mu}+w_{k_{j_3},\mu}) + \right. \\
&\left. + \sum_{j_1<j_2} K(w_{k_{j_1},\mu}+w_{k_{j_2},\mu}) - \sum_j K(w_{k_j,\mu})\right].
\end{align}
\end{subequations}
As can be seen from Eq.\ \eqref{I1k1}, the hidden layer makes an additive contribution to the bias term $b_{k_1}$ acting on node $k_1$ if there is at least one hidden node $\mu$ such that $w_{k_1,\mu} \neq 0$, noting that $K(0)=0$; similarly, the pairwise interaction, $I^{(2)}_{k_1,k_2}$, connecting nodes $k_1$ and $k_2$ is different from zero if and only if there is at least one hidden node $\mu$ such that $w_{k_1,\mu} \neq 0 $ and $w_{k_2,\mu}\neq 0$. In general, the {\em s}-body interaction connecting $s$ visible nodes can be different from zero if and only if there is at least one hidden node connecting the $s$ visible nodes, that is a hidden common input. It follows that the highest order of interaction among visible nodes that the RBM can represent corresponds to the largest number of visible nodes connected with the same hidden node. In particular, this suggests that, although, as we have seen in section \ref{sec:maci}, there is no choice of the potential $U(\bz)$ that leads to only a finite subset of interactions beyond pairwise models for every possible weight matrix $\bW$, it is still possible to limit the maximum order of interactions described by the RBM by constraining $\bW$ such that the maximum number of connections that each hidden node is bounded by some number smaller than $N$. Finally, it is clear from Eq.\ \eqref{eqn:general_int_terms} that an RBM can approximate any order of interaction among visible nodes with an accuracy determined by the choice of the potential and by the number of hidden nodes, becoming more accurate as the number of hidden nodes increases \citep{le2008representational}.

\begin{figure}[t]
\begin{center}
\centerline{\includegraphics[width=.7\columnwidth]{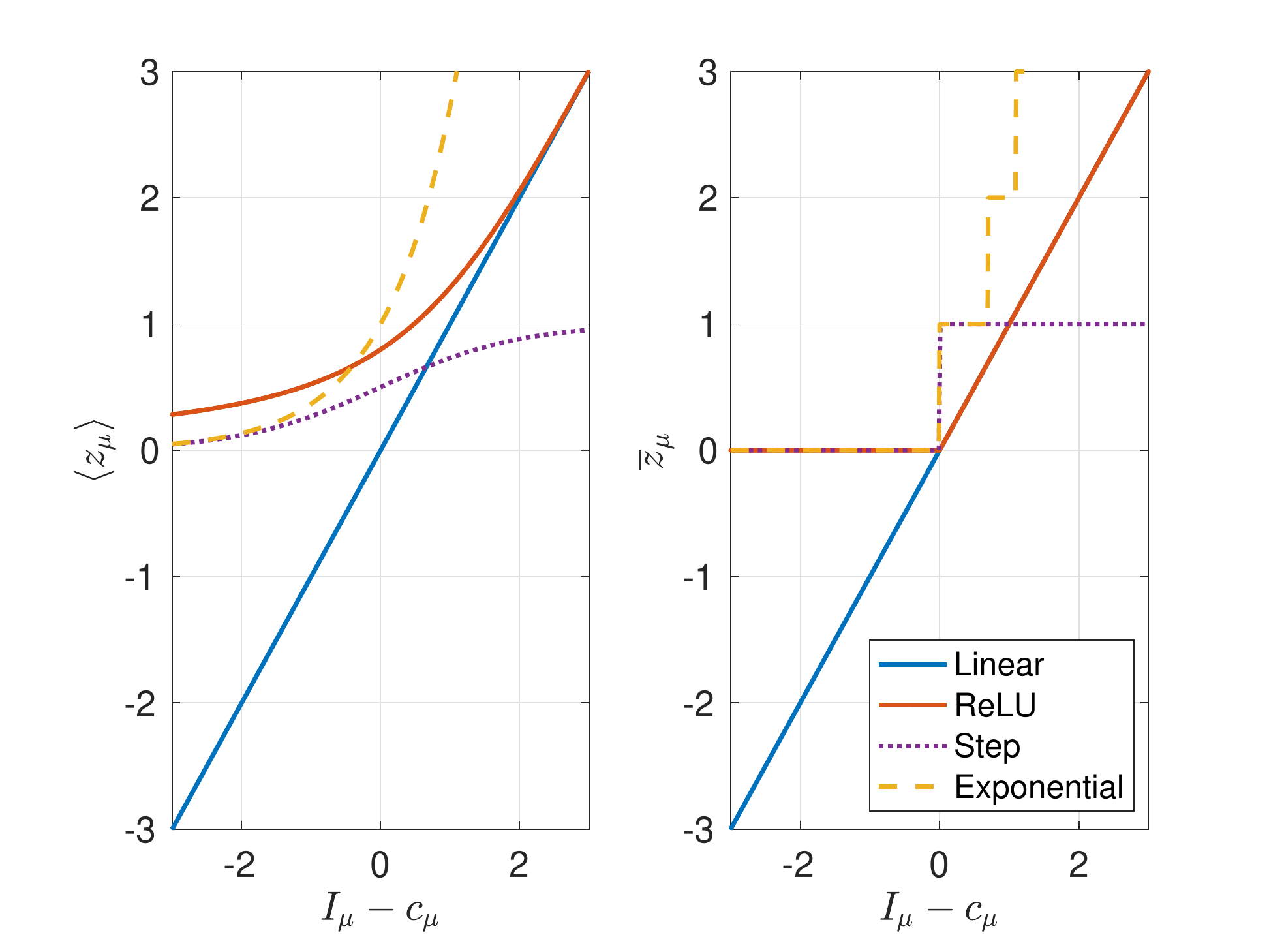}}
\caption{Mean activation (left) and mode (right) of the hidden node $z_\mu$ according to the conditional probability density $P(z_\mu|v)$ for each of the investigated potentials. The mode of the distribution represents the activation function.}
\label{activation_functions}
\end{center}
\end{figure}

The equivalence between RBMs and models of interacting binary variables allows one to compare on the same ground different RBMs and discuss their representational power by investigating how they parametrise the interaction parameters. In the case of separable potentials finding the interaction parameters in Eq.\ \eqref{eqn:general_int_terms_factorized} simplifies, as it requires calculating a one dimensional integral to find the generating functional
\begin{equation}\label{eqn:integral}
K(q) = \log\int dx \exp(x q) \rho(x),
\end{equation}
where $\rho(x) = \exp(-U(x))/\int dx \exp(-U(x))$, for each choice of the potential $U$. In the following sections we consider RBMs employing some of the most commonly used activation functions, i.e. Linear, ReLU, Step and Exponential activation functions. Note that the activation function is defined here as the mode of the conditional probability distribution of the hidden nodes given the visible nodes $P(\bz|\bv)$. The conditional probability is calculated by applying Bayes rule, $P(\bz|\bv) = P(\bv,\bz) / \int d\bz P(\bv,\bz)$. For a separable potential, where $U(\bz) = \sum_\mu U_\mu (z_\mu)$, the distribution factorises over hidden nodes and the probability of each hidden node conditioned on the visible activity is given by 
\begin{equation}\label{eqn:C_conditional_probability_separable}
P(z_\mu|\bv) = \frac{\exp\left[\bv\tr \bW_\mu z_\mu -U_\mu(z_\mu)\right]}{\int dz_\mu \exp\left[\bv\tr \bW_\mu z_\mu -U_\mu(z_\mu)\right]}.
\end{equation}
where $\bW_\mu$ is the $\mu$-th column of the matrix $\bW$.
The mode of the distribution, that is the value of $z_{\mu}$ at which the argument of the exponential in the numerator of Eq.\ \eqref{eqn:C_conditional_probability_separable} is maximised, can be expressed as $\overline{z}_\mu = {U'_\mu}^{-1} (I_\mu)$ where $ {U'_\mu}^{-1} $ is the inverse of the first derivative of $U_\mu$ and $I_\mu = \bv\tr \bW_\mu$ is the total input to the hidden node $\mu$ \citep{tubiana2017emergence}. The mode and the mean of the conditional distribution for the investigated potentials are shown in Figure~\ref{activation_functions}, while the exact expressions as well as their cumulant generating functions are reported in Appendix \ref{sec:Activationfunctions}. Note that $P(z_{\mu}|\bv)$ and its mode are both only functions of the visible nodes through $I_\mu$.

By substituting the expressions for the cumulant generating function $K_\mu(q_\mu)$ into Eq.\ \eqref{eqn:general_int_terms_factorized}, it is possible to evaluate the interaction terms for all previously considered cases. In this section, we discuss the differences and similarities in the way the resulting interaction terms are parametrised by the RBMs. 

We first explicitly write down the pairwise terms, that we denote by $J_{k_1,k_2}\equiv I^{(2)}_{k_1,k_2}$, for the different activation functions
\begin{subequations}\label{eqn:pairwise}
\begin{align}
J^{\mbox{\scriptsize Lin}}_{k_1,k_2}& = \sum_\mu w_{k_1,\mu}w_{k_2,\mu} \label{eqn:pairwiselin} \\
J^{\mbox{\scriptsize ReLU}}_{k_1,k_2} &= \sum_\mu w_{k_1,\mu}w_{k_2,\mu} + \log\frac{\left[1+\erf\left(\frac{w_{k_1,\mu}+w_{k_2,\mu}-c_\mu}{\sqrt{2}}\right)\right]\left[1+\erf\left(\frac{-c_\mu}{\sqrt{2}}\right)\right]}{\left[1+\erf\left(\frac{w_{k_1,\mu}-c_\mu}{\sqrt{2}}\right)\right]\left[1+\erf\left(\frac{w_{k_2,\mu}-c_\mu}{\sqrt{2}}\right)\right]} \label{eqn:pairwiseReLU}\\
J^{\mbox{\scriptsize Step}}_{k_1,k_2} &= \sum_\mu \log\frac{\left[1+\exp(w_{k_1,\mu}+w_{k_2,\mu}-c_\mu)\right]\left[1+\exp(-c_\mu)\right)}{\left[1+\exp(w_{k_1,\mu}-c_\mu)\right]\left[1+\exp(w_{k_2,\mu}-c_\mu)\right]} \\
J^{\mbox{\scriptsize Exp}}_{k_1,k_2} &= \sum_\mu \tilde\lambda_\mu u_{k_1,\mu} u_{k_2,\mu} \label{eqn:pairwiseexp}
\end{align}
\end{subequations}
where $u_{k,\mu} = \exp(w_{k,\mu}) -1$ and $\tilde\lambda_\mu = \exp(-c_\mu)$.

As can be seen in Eq.\ \eqref{eqn:pairwiselin} and noted by others \citep{martens2010parallelizable, barra2012equivalence}, the pairwise interaction matrix of an RBM with the Linear activation function has the same structure as the connectivity matrix in a Hopfield model. Interestingly, this Hopfield-like term arises also with any other activation function in the small $w$ limit. To show this, we first expand the cumulant generating function in Eq.\ \eqref{eqn:integral} for small $q$, as $K(q) = q \langle z \rangle + \frac{q^2}{2} \left( \langle z^2\rangle - \langle z\rangle^2\right) + O(q^3) $, where the averages $\langle \cdots \rangle$ are evaluated over the density $\rho(z)$. Plugging this expansion into Eq.\ \eqref{eqn:4int_terms_separable}, we find that to the leading order in $w$, pairwise interactions take the form of
\begin{equation}\label{eqn:small_w_approximation}
J_{i,j} = \sum_\mu \kappa_\mu^{(2)} w_{i,\mu}w_{j,\mu},
\end{equation}
where the differences among potentials are embedded in the second cumulant of the distribution, that is $\kappa_\mu^{(2)} = \langle z_\mu^2\rangle - \langle z_\mu\rangle^2 $. In the case of the Linear activation function, $ \kappa_\mu^{(2)} = 1$ and, indeed, $J^{\mbox{\scriptsize Lin}}_{i,j}= \sum_\mu w_{i,\mu}w_{j,\mu}$ in Eq.\ \eqref{eqn:pairwiselin}. Analogously, in the Exponential case where $\kappa_\mu^{(2)} = \tilde\lambda_\mu$, the small parameter approximation in Eq.\ \eqref{eqn:small_w_approximation} coincides with $J^{\mbox{\scriptsize Exp}}_{i,j} = \sum_\mu \tilde\lambda_\mu u_{i,\mu}u_{j,\mu}$ in Eq.\ \eqref{eqn:pairwiseexp} in the limit of small $w$ where $u_{k,\mu} \sim w_{k,\mu}$ . This relation to the Hopfield model makes learning in these RBMs more biologically plausible as it can be local. Moreover, small parameters are beneficial for training, especially for large networks and for quickly growing non-linear activation functions, as they prevent inputs from saturating the neural activity and, thus, leading to small gradients. In fact, we observe in our numerical Experiments on the MNIST dataset (see the next section) that Eq.\ \eqref{eqn:small_w_approximation} approximates quite well Eq.\ \eqref{eqn:pairwise} in almost all cases trained on this dataset, except when the RBM involves only a small number of hidden units, that is $M=50$ and $M=100$, where typically RBMs parameters are larger. 

We also report the expression for the higher order terms, i.e. those with $s \geq 3$, 
\begin{subequations}\label{eqn:ho_terms}
\begin{align}
I_{k_1,...,k_s}^{\mbox{\scriptsize Lin}} &= 0 \label{Linearho}\\
I_{k_1,...,k_s}^{\mbox{\scriptsize ReLU}} &=  \sum_{\mu = 1,p=0}^{M,s-1} (-1)^{p} \!\!\!\! \sum_{j_1< ... <j_{s-p}=1}^s \!\!\!\! \log\frac{1+\erf\left(\frac{\sum_{l = 1}^{s-p} w_{k_{j_l},\mu}-c_\mu}{\sqrt{2}}\right)}{1+\erf\left(\frac{-c_\mu}{\sqrt{2}}\right)} \label{ReLUho} \\
I_{k_1,...,k_s}^{\mbox{\scriptsize Step}} &= \sum_{\mu = 1,p=0}^{M,s-1} (-1)^{p} \!\!\!\!\sum_{j_1< ... <j_{s-p}=1}^s \!\!\!\!\log\frac{1+\exp(\sum_{l = 1}^{s-p} w_{k_{j_l},\mu}-c_\mu)}{1+\exp(-c_\mu)} \label{Stepho}\\
I_{k_1,...,k_s}^{\mbox{\scriptsize Exp}} &=  \sum_\mu \tilde\lambda_\mu \prod_{j=1}^s u_{k_j,\mu}. \label{Expho}
\end{align}
\end{subequations}
Details on the derivation of the interaction terms for the Linear and the Exponential activation function can be found in the Supplemental Material, whereas, for the other cases, results are straightforward. As before, by exploiting the expansion of the cumulant generating function, that is $K(q) = \sum_{n=1} \kappa^{(n)} q^n/n!$ , it is possible to derive the leading term of the small parameter expansion in $w$ for interactions of order $s$ that is
\begin{equation}\label{eqn:ho_small_w_approximation}
I_{k_1,...,k_s}^{(s)} = \sum_\mu \kappa_\mu^{(s)} w_{k_1,\mu}\cdots w_{k_s,\mu}.
\end{equation}
Notice that the above equation reduces to Eq.\ \eqref{eqn:small_w_approximation} for $s=2$ and that it presents a structure which is exactly the same as that of the Exponential activation function in Eq.\ \eqref{Expho} if we approximate $u = \exp(w) -1$ by its leading term as $w\to 0$, that is $u\approx w$ and noting that $\kappa^{(s)}_\mu = \tilde\lambda_\mu$, $\forall s$.

\subsection{Linear activation function and pairwise models}
While in general $I_{k_1,...,k_s} \neq 0$, $\forall s$, as discussed in section \ref{sec:maci}, the Linear activation function can only express pairwise models, i.e. $I^{\mbox{\scriptsize Lin}}_{k_1,...,k_s} \neq 0$ only for $s =1$ and $s=2$. As we argue below, an RBM with a linear activation function can represent any pairwise distribution over the observed nodes, either exactly or with the best possible approximation in the Frobenius norm depending on the number of hidden nodes.

Let us assume $M=N-1$ and suppose that we would like to have an RBM that represents a pairwise distribution over the visible nodes with interactions $\bJ$. Since any real symmetric $\bJ$ can be decomposed as $\bJ = \bU \bLambda \bU\tr$, where $\bU$ is an orthogonal matrix and $\bLambda$ is the diagonal matrix of the eigenvalues of $\bJ$ which are all real, it follows that $\bJ$ can be written as
\begin{equation}\label{eqn:jfromw}
\bJ = \bU \bLambda \bU\tr = \bU (\bLambda + \lambda_0 \bm I) \bU\tr - \lambda_0\bm I = \bB\bB\tr - \lambda_0\bm I 
\end{equation}
where $\bm I$ is the identity matrix and $\lambda_0$ is the smallest eigenvalue of the matrix $\bJ$. The matrix $\bB = \bU\sqrt{\bLambda + \lambda_0\bm I}$ is a square matrix of size $N$ with one column having all entries equal to zero, since $\bLambda + \lambda_0\bm I$ is a diagonal matrix with one diagonal element equal to zero\footnote{We consider for simplicity the case in which the minimum eigenvalue is not degenerate, otherwise the number of zero diagonal elements is equal to the degree of degeneracy of that eigenvalue.}. Unless the minimum eigenvalue is degenerate, the matrix $\bB\bB\tr$ is of maximum rank $N-1$. Noting from Eq.\ \eqref{eqn:pairwiselin} that $\bJ^{\mbox{\scriptsize Lin}}= \bW\bW\tr$, in order to construct an RBM with a linear activation function that represents pairwise interactions between visible nodes given by the matrix $\bJ$, we can therefore select the hidden-visible interaction matrix $\bW$ to be equal to $\bB$ and include $\lambda_0$ in the bias term.

When $M < N-1$, in general, the representation $\bJ^{\mbox{\scriptsize Lin}} =\bW\bW\tr$ cannot match exactly the matrix $\bB\bB\tr$. In this case, the RBM approximates it with a lower rank matrix of order $M+1$. In particular, the approximation can partly reproduce the eigenvalue decomposition of $\bB\bB\tr$ by matching the largest $M$ eigenvalues and thus achieving the best approximation in the Frobenius norm of the distance between the matrix and its approximation. In fact, more generally, Eckart-Young theorem ensures that the best low-rank approximation in terms of this Frobenius norm distance is given by the truncated singular value decomposition \citep{eckart1936approximation}. This means that the parametrisation $\bJ^{\mbox{\scriptsize Lin}}$ in Eq.\ \eqref{eqn:pairwiselin} provides the most efficient representation of pairwise interactions, namely the most accurate for a fixed number of hidden nodes.

\subsection{ReLU activation function, pairwise models and beyond}
The parametrisation of pairwise interactions for the ReLU activation function in Eq.\ \eqref{eqn:pairwiseReLU} consists of two terms, one of which is the Hopfield-like term, $\bW\bW\tr$, which we already encountered in the linear case. The parameter vector $\bc$,  which acts as a threshold for the ReLU, does not affect the Hopfield-like term, but it affects the value of the second term in Eq.\ \eqref{eqn:pairwiseReLU}. This is true also for all higher order terms which are regulated by $\bc$ (see also Eq.\ \ref{ReLUho}). For example, it is easy to verify from Eq.\ \eqref{eqn:pairwise} that the Linear case would be retrieved for large negative values of the elements of $\bc$ and randomly distributed weights with a variance that scales as $1/N$ or for negative values of $\bc$ and non-negative weights. As a result, the threshold vector $\bc$ allows the RBM to tune the relevance of higher order terms and the departure from the pure Hopfield-like representation by appropriately selecting $\bc$.  This is again consistent with intuition: in both examples the input to the hidden nodes is always above threshold, which is the regime where ReLU is effectively acting like a linear function.

The ReLU activation function is also interesting because it somehow lies between the cases of Linear and Step activation functions. In fact, it employs the Hopfield-like term for the pairwise interactions, like in the Linear case, and it expresses higher order terms in a way which is similar to the representation arising with a Step activation function (standard RBM): this can be seen from Eqs.\ \eqref{ReLUho} and \eqref{Stepho} by noting that one can choose a parameter $\beta$ such that $1+\erf(x) \approx  2/(1+\exp(\beta x))$ for that value of $\beta$, which is true with a good approximation at least for $x \gtrapprox -1$. Actually, from this perspective, an RBM with ReLU could be interpreted as a standard semi-Restricted Boltzmann Machine with visible-visible and visible-hidden connections parametrised differently by the same set of parameters.

\subsection{Exponential activation function}
\label{ExponentialAll}
Finally, the case of an Exponential activation function is particularly interesting because it employs an encoding procedure for the higher order terms which is the natural extension of the efficient pairwise representation discussed in the Linear case. In fact, this particular RBM with $M$ hidden nodes approximates, by using the same set of weight parameters, every interaction term with the closest possible approximation to its {\itshape symmetric outer product decomposition} of order $M$, which might be regarded as a generalisation of the eigenvalue decomposition to symmetric tensors \citep{comon2008symmetric}.

As a last remark, we note that in the Linear case, there are multiple solutions $\bW$ parametrising the same interaction matrix $\bJ^{\mbox{\scriptsize Lin}} =\bW\bW\tr$. For instance, it is known that the weight matrix is invariant under rotations of the parameter matrix $\bW$. This fact hinders the interpretation of the hidden node representation since it depends on the choice of $\bW$. In Factor Analysis, which can be considered the directed counterpart of the RBM, this problem is known as the ``unidentifiability" of the latent representation (see e.g. \citet{murphy2012machine}). Several tricks have been proposed to overcome this issue in applications involving Factor Analysis and RBMs \citep{welling2005exponential,murphy2012machine}. For the ReLU and the Exponential activation function cases, this symmetry is naturally broken by the presence of higher order interaction terms: since there are many set of parameters which are equally well describing pairwise interactions, the learning algorithm will select the set that best fits also the higher-order terms with the same set of parameters. This fact promotes the identifiability of hidden variables which is valuable in model interpretation and latent variable discovery.


\section{Numerical Results}\label{sec:numerical_experiments}

Following the theoretical discussions of the previous sections, in this section, we report our numerical results. We first report some simulations in which we compare the resulting interaction models over the binary visible nodes that emerge from our theoretical expressions compared to what we actually find from simulating simple RBMs. We then study the performance and the resulting interaction model when we train RBMs with different number of hidden nodes and activation functions on the MNIST data. 

\subsection{Numerical evidence of the mapping}\label{sec:numerical_verification}
\begin{figure}[t]
\begin{center}
\includegraphics[width=.8\columnwidth]{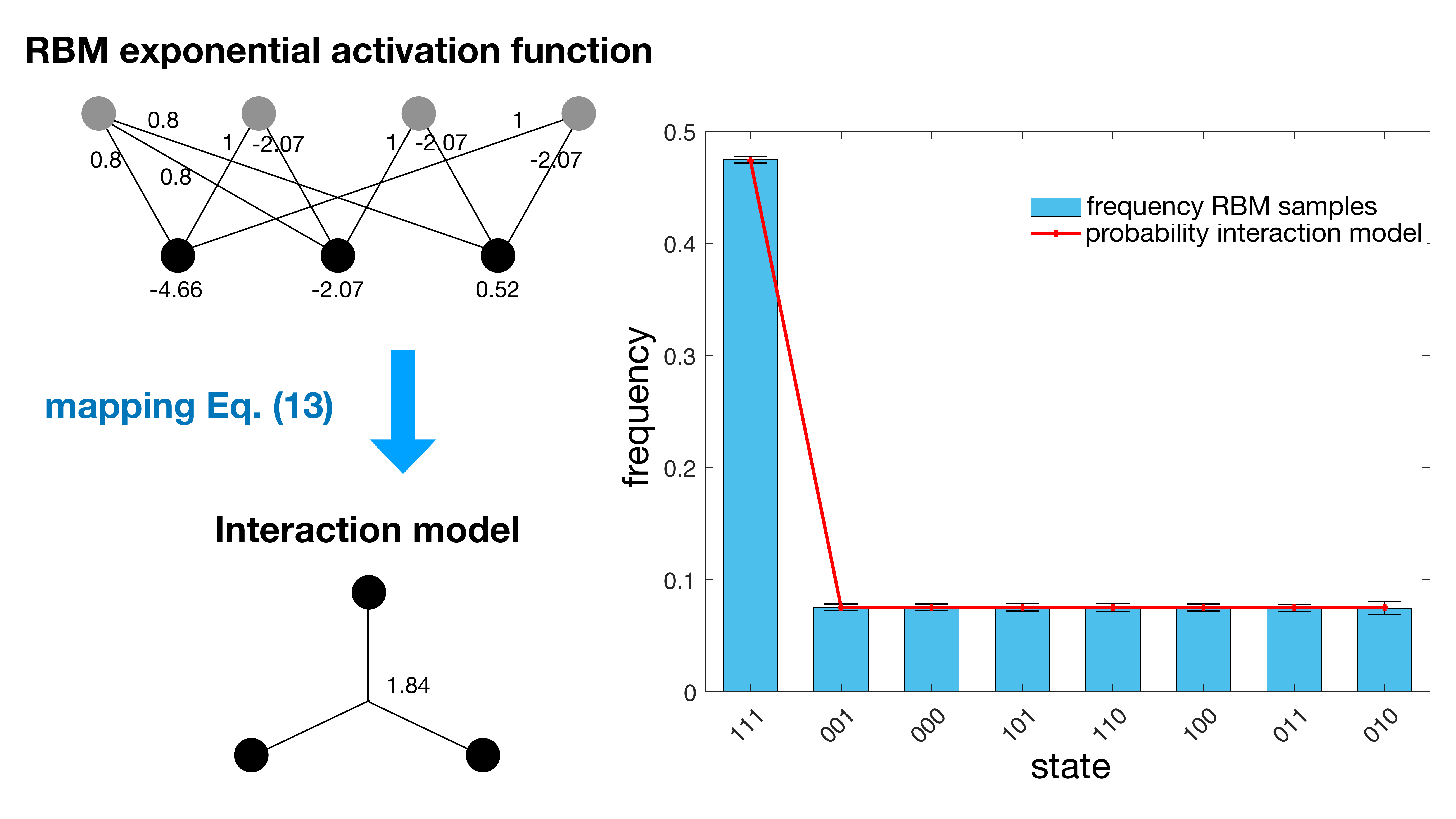}
\caption{Illustration of the mapping in a simple example of an RBM which is equivalent to an interaction model with only a three-body interaction term. The values of non zero parameters are reported in the figure for both the RBM and the interaction model. The frequencies of observation of the 8 possible states are measured in samples of 1000 data-points each drawn from the RBM. The histogram in the figure shows the mean and standard deviation of the measured frequencies over 20 independent samples (cyan bars) and the probabilities of the same states as calculated with the equivalent interaction model (red line).}
\label{fig:RBMexample1}
\end{center}
\end{figure}

In this first numerical experiment, we consider a simple RBM with $N=3$ visible nodes, $M=4$ hidden nodes and the Exponential activation function as shown in Figure~\ref{fig:RBMexample1}. The weights that are not drawn in the figure, as well as the biases on the hidden units, are set to zero. By applying the mapping described in the previous section, namely Eq.\ \eqref{eqn:general_int_terms_factorized}, it is easy to verify that this RBM is equivalent to a pure three-body interaction model, that is a model in which the three visible nodes are interacting among each other only through a three-body interaction term. The value of the three-body term is also reported in the figure. The equivalence between the RBM and the corresponding interaction model is proved by sampling configurations from the RBM with blocked Gibbs sampling, measuring the frequency of observation of each state and comparing those frequencies with the probabilities as evaluated with the interaction model. This is shown in the histogram in Figure~\ref{fig:RBMexample1}: the frequencies by which each state appears in the samples generated by the RBM (cyan bars) match the probabilities of the same state as predicted by the equivalent interaction model (red line).

\begin{figure}[t!]
\begin{center}
\includegraphics[width=.8\columnwidth]{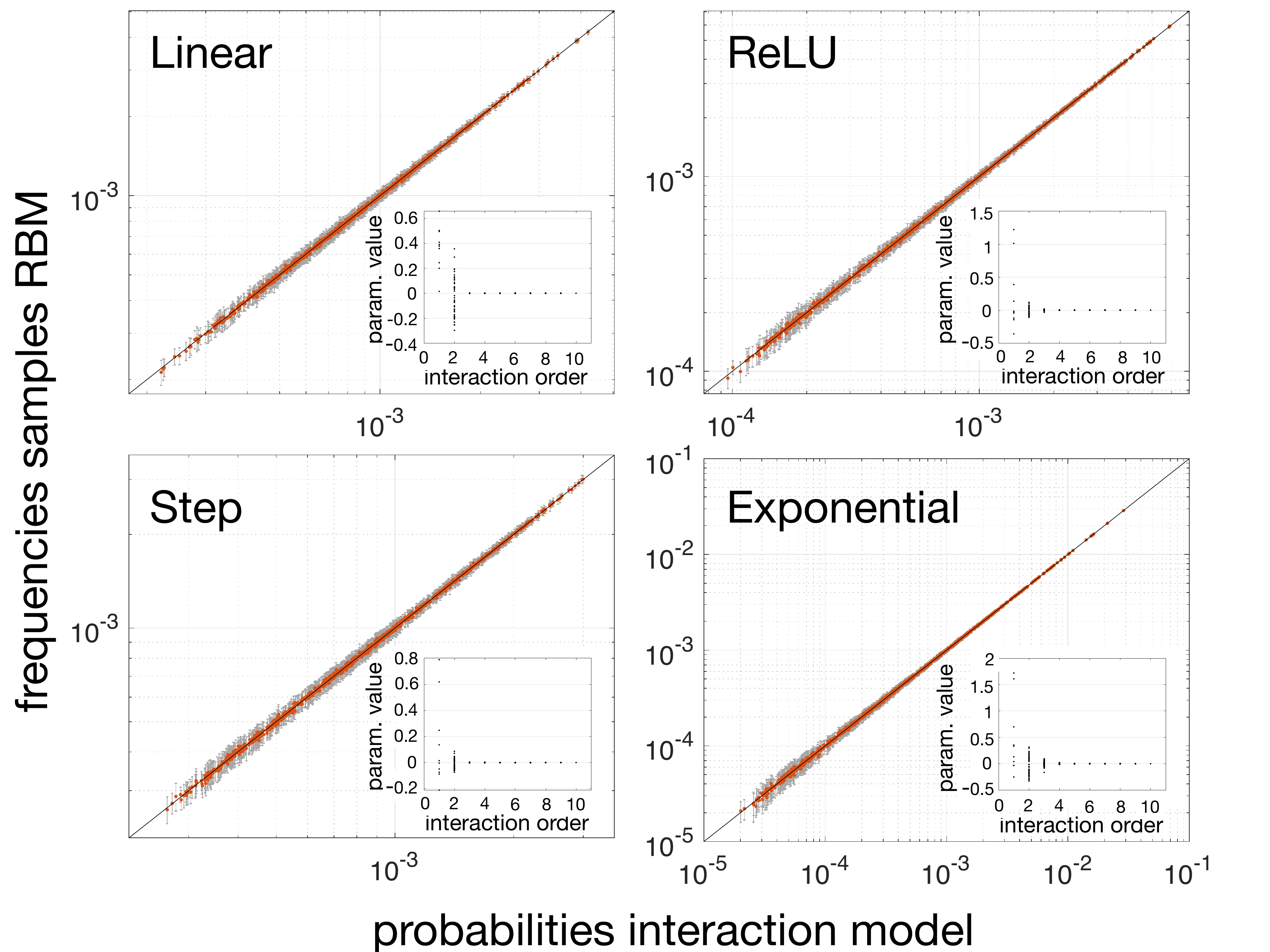}
\caption{Examples of RBMs with random parameters drawn from a gaussian distribution with zero mean and standard deviations equal to $1/\sqrt{M}$ for the weight parameters and 0.1 for the bias parameters. For each of the four activation functions discussed in the previous section, an RBM is defined as described above and a sample of 512000 data-points (500 times the size of the state space) is drawn from it. The figures exhibit a good alignment between the mean frequency of observation of states across 20 independent samples and the probabilities of the same states as calculated with the equivalent interaction model. Error bars correspond to standard deviation of frequency values across the trials. Insets of the figures report parameter values of the corresponding interaction model as a function of the interaction order.}
\label{fig:RBMexample2}
\end{center}
\end{figure}

In the second numerical experiment, reported in Figure~\ref{fig:RBMexample2}, we checked the analytical mapping of the previous section by considering larger RBMs with $N = 10$ visible and $M=15$ hidden nodes. The parameters of these RBMs are chosen randomly: the weights are drawn from a Gaussian distribution with zero mean and standard deviation $1/\sqrt{M}$, whereas the values of the hidden and visible biases are drawn from a Gaussian distribution with zero mean and standard deviation equal to 0.1. This choice for the parameters is to ensure that the resulting distribution in the state space is not very peaked so that a good number of observations for the majority of the states is expected. The values of parameters in the corresponding interaction model are evaluated through the mapping and shown in the insets of the figure as a function of the interaction order. Notice that in the Linear case all interactions of order larger than two are equal to zero, consistent with the analytical results of the previous sections. Samples of 512000 data-points (500 times the size of the state space) are drawn from RBMs with all the four activation functions examined in the previous section. Scatter plots in Figure~\ref{fig:RBMexample2} show that frequencies of the states observed in the samples are well aligned with probabilities as evaluated with the equivalent interaction model.

As reported in Eq.\ \eqref{eqn:ho_small_w_approximation}, if the typical value of the weight strength is $w$ then to the leading order in the small parameter expansion of the interaction terms, a $s$ body interaction term is proportional to $w^s$, that is $I^{(s)} \sim O(w^s)$. Consequently, if $w$ is small, the strength of higher order terms decays quite rapidly with the interaction order. In fact, in the experiment illustrated in Figure~\ref{fig:RBMexample2}, where weights are normally distributed with a standard deviation that ensures that the magnitude of typical inputs to the hidden units are smaller than one, the RBMs are very well represented by pairwise models.

\begin{figure}[t!]
\begin{center}
\includegraphics[width=\columnwidth]{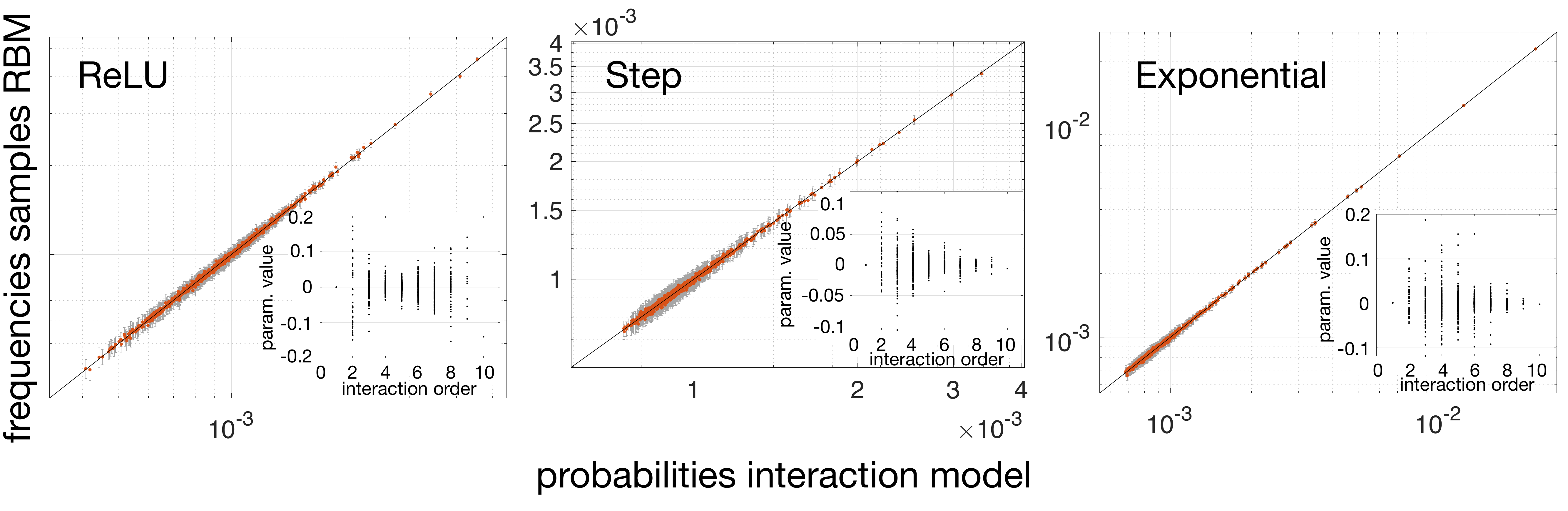}
\caption{Examples of RBMs with random weights drawn from a gaussian distribution with zero mean and unit standard deviations. Hidden biases are all set to $c=5$, whereas visible biases $b$ are selected so that corresponding interaction models would have first order interaction terms equal to zero. For each of the three non-linear activation functions, an RBM is defined as described above and a sample of 512000 data-points (500 times the size of the state space) is drawn from it. The figures exhibit a good alignment between the mean frequency by which the states appear in the data across 20 independent samples and the probabilities of the same states as calculated with the equivalent interaction model. Error bars correspond to standard deviation of frequency values across the trials. Insets of the figures report parameter values of the corresponding interaction model as a function of the interaction order.
}
\label{fig:RBMexample3}
\end{center}
\end{figure}

One way to obtain a random RBM that represents an interaction model with relatively high values of higher order interaction terms is to increase the values of the weights and those of the hidden node biases, that is by making visible-hidden connections stronger and, at the same time, hidden nodes more sparsely activated. Accordingly, in our third experiment, we consider RBMs with random weights drawn from a Gaussian distribution with zero mean and unit variance and with larger values for the hidden unit bias ($c=5$ for all hidden units). By setting $I^{(1)} = 0$ in Eq.\ \eqref{eqn:4int_terms_separable} and solving for $b$, we fix the values of the visible node biases such that the first order interaction terms have values equal to zero. Figure~\ref{fig:RBMexample3} reports the results of these numerical experiments for the three non-linear activation functions showing, as before, a good agreement between mean frequencies of states as observed in the samples and probabilities calculated with the interaction model. Moreover, by looking at the insets in the figure, it is clear that the interaction models generated in this way exhibit relevant higher order interaction terms. 

\subsection{RBMs on MNIST}
We now turn our analysis on studying the properties of the RBMs fitted to the MNIST data.
The MNIST dataset is subdivided into a training set containing 60000 images and a test set with 10000 images. Each image is composed of 28-by-28 pixels representing a digit from 0 to 9 and each pixel can take values between 0 and 255. In order to use RBMs as generative models, data is binarised by thresholding pixel intensity at the midpoint value, that is by transforming each pixel into a binary variable which can take values of 0 or 1 according to whether its intensity is smaller or larger than 128 respectively. After that, images are flattened into a binary vector of $N = 784$ entries and training samples are fed to the learning algorithm.

Training is performed by employing the Contrastive Divergence algorithm with mini-batches of 100 data-points \citep{hinton2002training}. An adaptive learning rate and an increasing number of steps in the CD-k algorithm are used during training \citep{salakhutdinov2008quantitative}. In particular, the number of steps increases with the number of epochs as $k = \mbox{ceil}(\mbox{epoch}/10)$ while the learning rate decreases as $\eta = \eta_0/k$. Thus, by the end of training, learning proceeds as in a CD-50 scheme with learning rate $\eta = \eta_0/50$. The initial value $\eta_0$ for the learning rate is chosen differently for each activation function and as small as possible in order to ensure a smooth convergence within 500 epochs (300000 updates in total) for any choice of the hidden layer size. At the beginning of training, visible biases are initialised as $\bb = \log(\langle \bv \rangle)-\log(1-\langle \bv \rangle)$ where $\langle \bv \rangle$ is the average over the entire training set and where a very small pseudo-count is used to avoid null entries in the $\langle \bv \rangle$ vector. Each weight parameter is initialised as $\sqrt{0.1/N}$ with a random sign, whereas the hidden biases are initialised as $\bc = \bW\langle \bv \rangle$ so that the input to each hidden node is distributed around zero. This is to discourage the formation of big inputs which could cause divergences during training, especially in the case of unbounded and strictly positive activation functions such as the ReLU and the Exponential ones.

\subsubsection{Performance of the RBMs with different activation functions on MNIST data}
\begin{figure}[t]
\begin{center}
\includegraphics[width=\columnwidth]{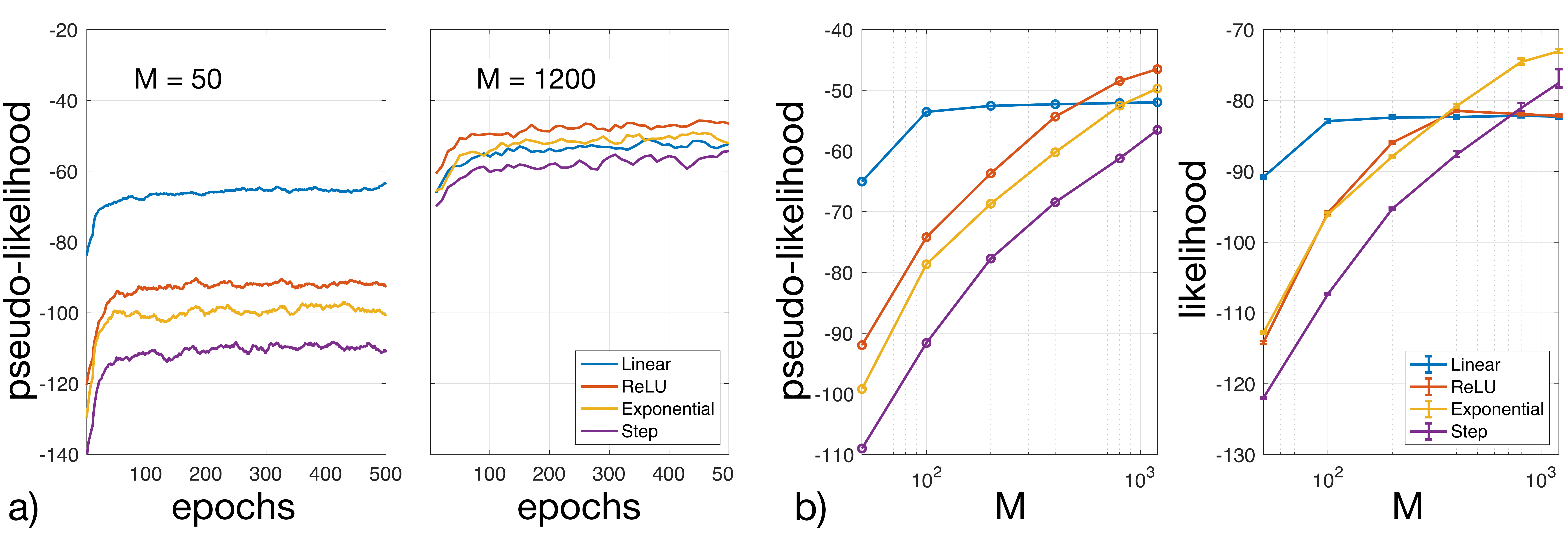}
\caption{a) Pseudo-likelihood evaluated from random subset of 100 data-points drawn from the test set during learning and averaged with a moving window of 20 epochs for the cases of M = 50 and M = 1200 hidden nodes; b) Pseudo-likelihood and likelihood of the trained models evaluated over the entire test set as a function of the number of hidden nodes employed by the RBM.}
\label{MNIST_results}
\end{center}
\end{figure}

Learning is monitored via the pseudo-likelihood function evaluated after each parameter update over a random subset of 100 data-points drawn from the test set. Learning curves are shown in Figure~\ref{MNIST_results}a for RBMs with 50 and 1200 hidden nodes and for all activation functions studied in the previous sections. Values of the pseudo-likelihood evaluated over mini-batches of the training set resemble very closely those reported in the figure for the test set and, therefore, they are not showed. From the figure it is clear that, when 50 hidden units are employed, the pseudo-likelihood values of the RBM with a Linear activation function are always larger than those of the other RBMs with non-linear activation function, whereas, except for the Step activation function, the converse is true for 1200 hidden units. This fact is evident in Figure~\ref{MNIST_results}b, where the pseudo-likelihood and likelihood values, evaluated on the entire test set, are shown as a function of the number of hidden nodes and by varying the activation function. The likelihood was estimated with the Annealing Importance Sampling (AIS) procedure, employing 100 annealing runs and around 14000 intermediate temperatures \citep{neal2001annealed}. Following \citet{salakhutdinov2008quantitative}, error bars around the likelihood estimates are computed from the predicted bounds on the estimates of the partition function. The upper (lower) bound is determined as the mean plus (minus) three times the standard error of the mean over the annealing runs. Annealing runs returning values larger than three times the scaled median absolute deviations (MAD) from the median, as detected by the ``rmoutliers" MATLAB routine, were considered outliers and removed from the computation of the mean and error bars. These few atypical runs (two or three runs on average and occurring only in few cases) were clearly unreasonable estimates of the partition function and most likely due to the Gibbs sampler getting stuck in some local minima.

Figure~\ref{MNIST_results}b shows that the RBM with the Linear activation function does not benefit from increasing the size of the hidden layer beyond $M = 100$: the values of both the pseudo-likelihood and the likelihood saturate after that critical size, which reflects the fact that this RBM is limited to capturing only pairwise interactions and augmenting the hidden layer size can only increase the accuracy with which these interactions are represented. Eventually, the figure demonstrates that pairwise interactions are already well captured by a matrix of rank around 100. 

For small hidden layer sizes ($M \lesssim 400$), the RBM with the Linear activation function achieve the largest values of both the pseudo-likelihood and the likelihood. In light of the findings of the previous section, this observation could be due to the fact that few parameters are enough to describe accurately the pairwise correlations, which is what the Linear activation function does (see Eq.\ \ref{eqn:pairwiselin}), while they are insufficient to represent at the same time and with a good approximation higher order correlations. Moreover, this observation also points to the fact that a large portion of the variability in the data is well captured by pairwise interactions which are efficiently described by the RBM with the Linear activation function.

Non-linear activation functions instead are not limited to modelling only the pairwise interactions between visible nodes and so their ability to describe the data increases steadily with the number of hidden nodes. Values of the likelihood for the standard RBM (the one employing a Step activation function) are comparable with those reported in other studies \citep{martens2010parallelizable,salakhutdinov2008quantitative} and are always smaller than those obtained with the other RBMs for $M \leq 400$. Surprisingly, the likelihood of the RBM with ReLU saturates after $M=400$, similarly to the RBM employing the Linear activation function. This fact is unexpected given the higher representational power of RBMs with ReLU and it might be a consequence of the learning procedure: although a non-linear activation function allows the machine to capture details beyond pairwise correlations and a large number of hidden nodes provides the necessary resources (parameters) to achieve this goal, the learning algorithm might fail to explore relevant configurations in the parameter space. In fact, more generally, learning algorithms might be biased towards certain regions of the parameter space \citep{baldassi2020shaping}. 

Finally, RBMs with the Exponential activation function exhibits likelihood values larger that any other RBMs for $M>400$. This efficiency can likely be attributed to the fact that, as discussed in section \ref{ExponentialAll}, this type of RBS effectively approximate the symmetric tensor decomposition of the interactions of all orders. 

\subsubsection{The effective interactions between the visible nodes}

In this section, we explore in more detail the properties of the resulting interaction parameters that we find from the expression in Eq.\ \eqref{eqn:general_int_terms_factorized} for the different activation functions. In particular, we investigate how the resulting parameters depend on the number of hidden nodes and on the choice of the activation function.

\begin{figure}[t]
\begin{center}
\includegraphics[width=\columnwidth]{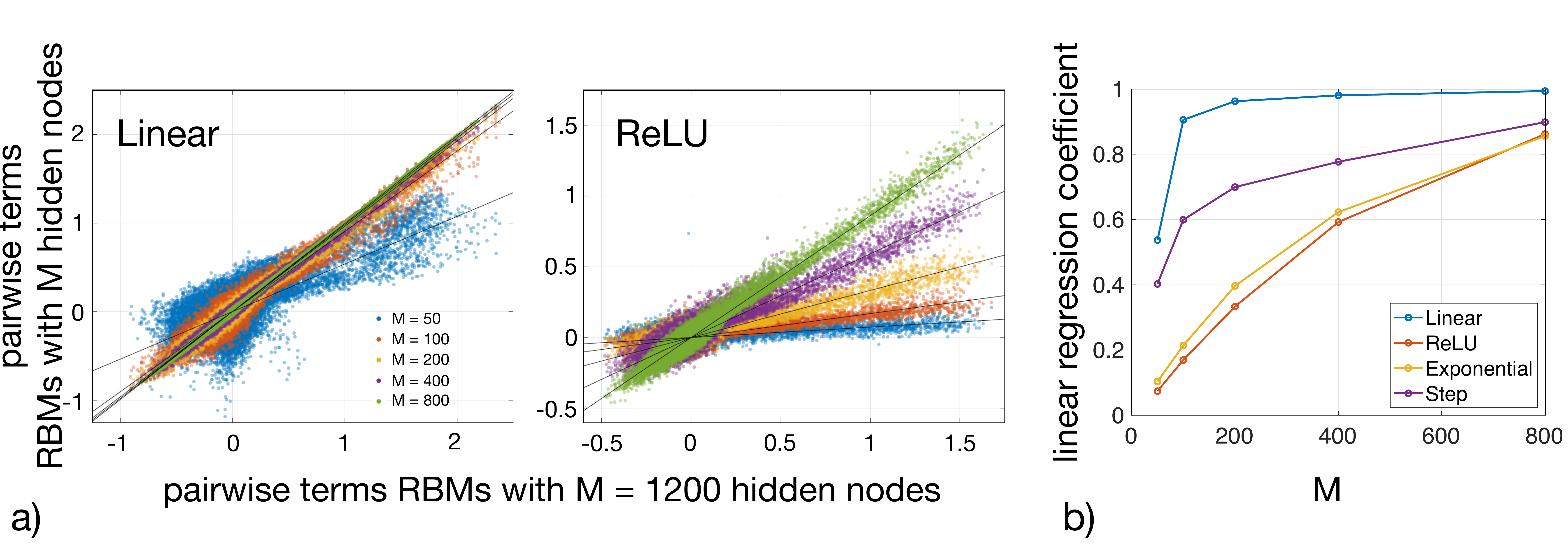}
\caption{a) Pairwise interaction terms induced by RBMs with different hidden layer sizes as compared to those induced by RBMs with M = 1200 hidden nodes for the Linear and ReLU activation functions. b) Linear regression coefficients showing how the slopes of the regression lines displayed in a) change with the number of hidden nodes M employed by the RBMs.}
\label{fig:pairwiseMNISTvsM}
\end{center}
\end{figure}

In Figure~\ref{fig:pairwiseMNISTvsM}, we show how the pairwise interactions change as we vary the number of hidden nodes $M$. In the case of the Linear activation function (Figure~\ref{fig:pairwiseMNISTvsM}a), pairwise interactions do not change substantially after $M = 100$, but only get refined by the larger number of hidden nodes. The same is true also for the resulting bias terms (not shown), which we will refer to as the field parameters. Given that fields and pairwise interactions are the only parameters that an RBM with the Linear activation function can express, the fact that these parameter do not change substantially after $M = 100$ is consistent with the saturation of the likelihood and pseudo-likelihood functions shown in Figure~\ref{MNIST_results}b. In the case of non-linear activation functions, such as the ReLU in Figure~\ref{fig:pairwiseMNISTvsM}a, the magnitude of pairwise interactions increases with $M$. Again, this is true also for the field parameters. Besides getting stronger, there appears to be a clear structure in the way the pairwise interactions in the non-linear cases change with $M$: the most important effect of the number of hidden nodes is a simple rotation in the scatter plots, namely a linear scaling with $M$. We have estimated the scaling parameter for the different activation functions by regressing the pairwise interaction matrix $J$ for a given $M$ against a reference $J$, taken to be as that with $M=1200$, namely $J^{M} = \alpha_{M} J^{M=1200}$. The results are plotted in Figure~\ref{fig:pairwiseMNISTvsM}b from which we can see that $\alpha_{M}$ grows large as $M$ approaches $1200$, but it happens at different rates for different activation function: for the Linear case, it increases rapidly reaching $0.9$ already at $M=100$, consistent with Figure~\ref{fig:pairwiseMNISTvsM}a, while the slope is slowest for the ReLU.

\begin{figure}[t]
\begin{center}
\includegraphics[width=.8\columnwidth]{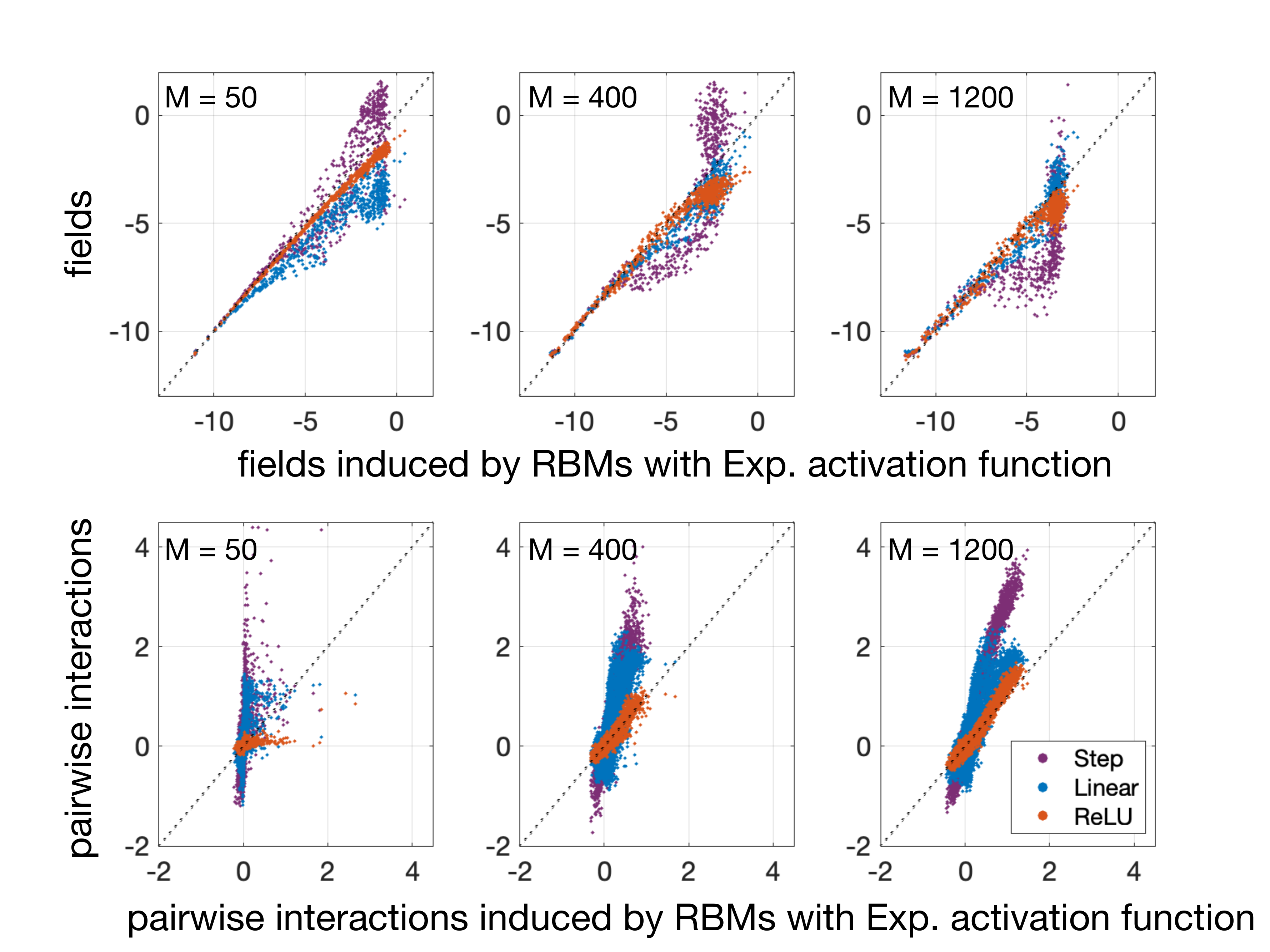}
\caption{Fields (first row) and pairwise interactions (second row) induced by RBMs with different activation functions against the same parameters as retrieved by employing RBMs with an Exponential one for three values of the hidden layer size, i.e. $M=50$, $M= 400$ and $M = 1200$.}
\label{fig:fieldsandcouplingsvseachother}
\end{center}
\end{figure}

After studying the dependence of the pairwise interactions on the number of hidden nodes, we wondered how the fields and pairwise interactions induced on the visible units in the RBMs with different activation functions relate to each other. The results are shown in Figure~\ref{fig:fieldsandcouplingsvseachother} for the fields and the pairwise interactions. As can be seen from this figure, fields induced by the RBMs with Exponential and ReLU activation functions are very similar to each other. The other activation functions show larger differences when compared to the Exponential activation function and across each other, more so the Step function and less so the Linear. Large negative values of the fields are all equal across activation functions which is not surprising as these are attached to pixels in the MNIST images that are off very often (fields with values around $-30$, corresponding to pixels almost always off, are not shown, but they are also consistent across activation functions). Similarly, in the case of pairwise interactions, parameters induced by the Exponential and ReLU activation functions align quite well for $M \gtrsim 400$. In the same range of $M$, pairwise interactions induced by the Step function seems to be related to those induced by the Exponential function (and, by extension, also to those induced by the ReLU function) through a simple linear scaling. Finally, pairwise interactions described by the Linear RBM also appear to show similarities with the corresponding parameters arising from non-linear activation functions, although there is much larger variability. 

One can also observe from Figure~\ref{fig:fieldsandcouplingsvseachother} that in general, regardless of $M$, fields are in better agreement with each other than the pairwise interactions are. This difference is even more pronounced when we look at the induced third order interaction in Figure~\ref{fig:three_body}. This observation can be explained by observing that as per Eq.\ \eqref{I1k1}, there are two terms that contribute to the fields. The first one, $b_{k_1}$ is independent of the form of the activation function and corresponds to the bias term, while the second one depends on the activation function through $K$. If the  parameters $w$ are small, this will make the relevance of first term more substantial, which being independent of the form of the activation function leads to the larger similarity of induced fields across activation function. On the other hand, as we show below, higher order parameters are more susceptible to the statistical error of RBM parameter estimation which contributes to differences among activation functions for higher order terms.

In summary, each non-linear activation functions returns values for the pairwise interactions which are related to each other by a linear scaling, across different choices for the hidden layer size $M$ (Figure~\ref{fig:pairwiseMNISTvsM}). The same is true for a given $M$ and across different non-linear activation functions, at least for $M \gtrsim 400$ (Figure~\ref{fig:fieldsandcouplingsvseachother}).

\begin{figure}
\begin{center}
\includegraphics[width=.7\columnwidth]{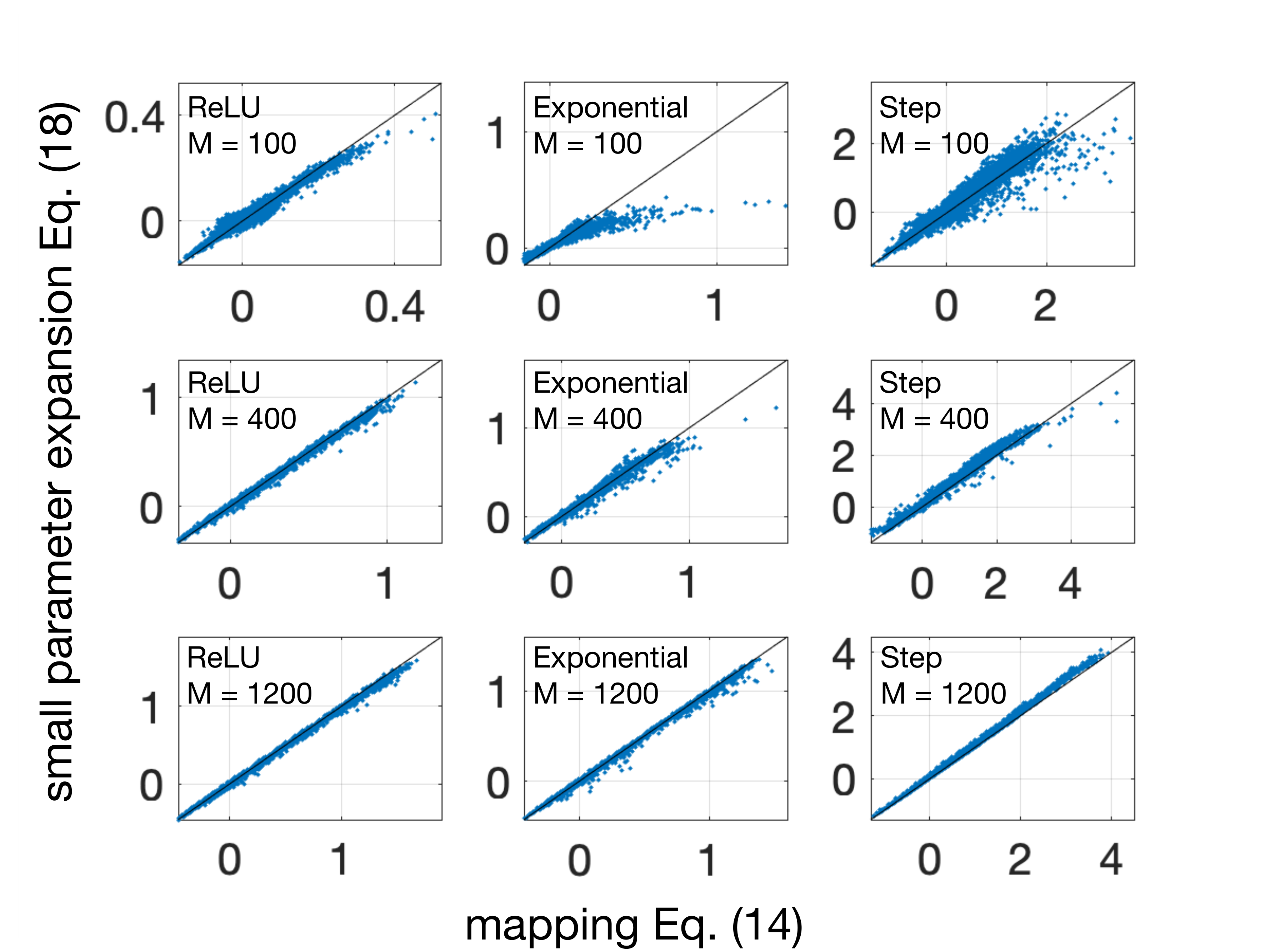}
\caption{Pairwise interactions induced on the visible nodes as approximated by Eq.\ \eqref{eqn:small_w_approximation} versus their exact values as evaluated with Eq.\ \eqref{eqn:general_int_terms_factorized} for the three non-linear activation function investigated and for $M = 100, 400, 1200$.}
\label{fig:pairwiseMNISTsmallweights}
\end{center}
\end{figure}

\begin{figure}[t]
\begin{center}
\includegraphics[width=.7\columnwidth]{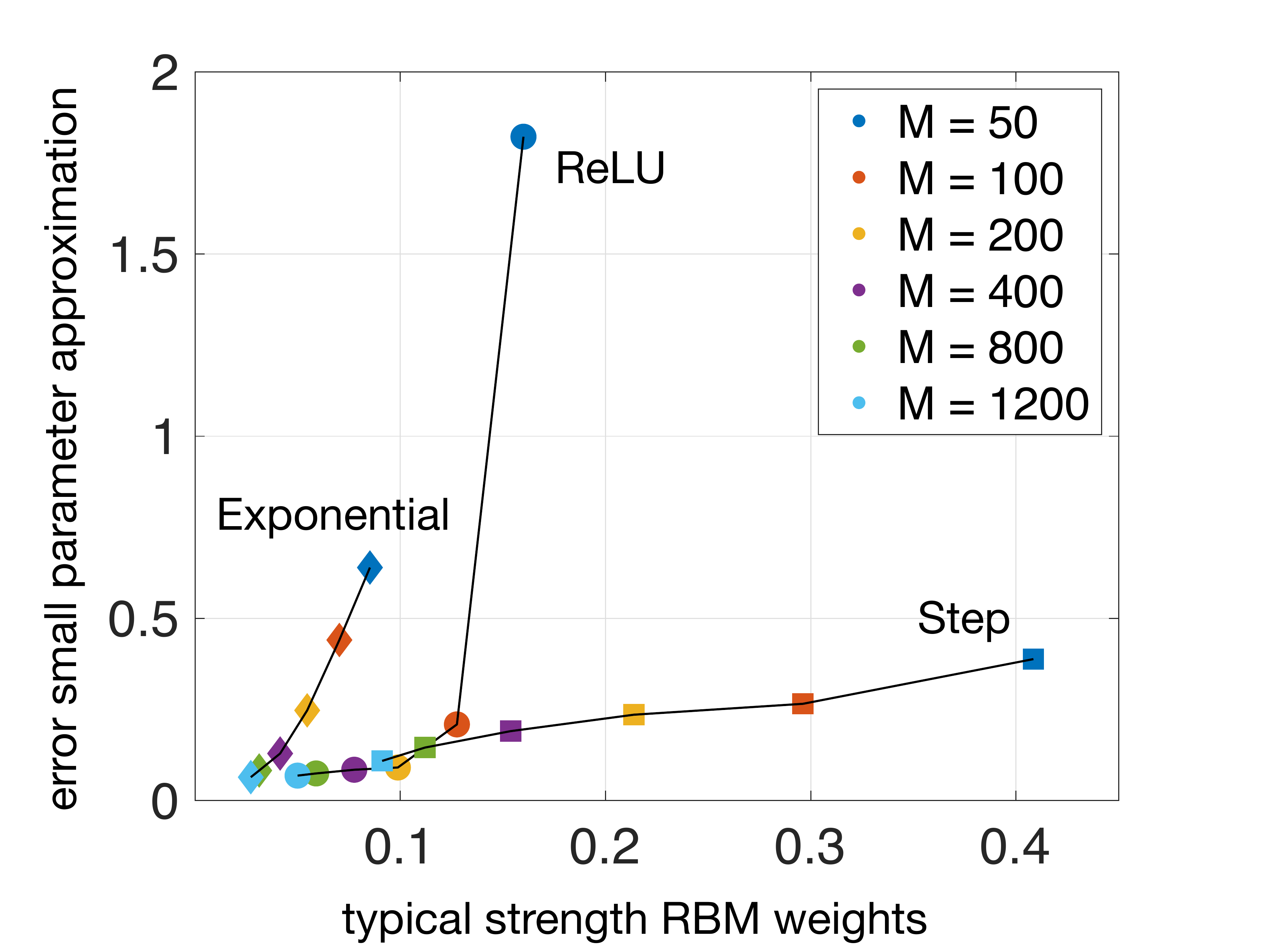}
\caption{Root mean square error between the weak parameter approximation to the pairwise interactions, Eq.\ \eqref{eqn:small_w_approximation}, and the values computed using Eq.\ \eqref{eqn:pairwise} from the trained weights vs. the root mean square of the trained weights RBM for the Exponential (diamonds), ReLU (circles) and Step (squares) activation functions and for different values of $M$ (color coded).}
\label{fig:pairwiseMNISTsmallweights2}
\end{center}
\end{figure}

As discussed in section \ref{sec:seppotential}, in the limit of small RBM weights $w_{i,\mu}$, the pairwise interactions induced on the visible nodes can be approximated by Eq.\ \eqref{eqn:small_w_approximation}. In order to assess the quality of the approximation on the MNIST dataset, we compare the pairwise interactions that are found by applying the mapping of Eq.\ \eqref{eqn:general_int_terms_factorized} with the same interactions as calculated from Eq.\ \eqref{eqn:small_w_approximation} for different activation functions and number of hidden nodes. The expressions for the second cumulant, which are needed for evaluating Eq.\ \eqref{eqn:small_w_approximation}, are reported in the Appendix for the investigated activation functions. The results of the comparison are shown as scatter plots in Figure~\ref{fig:pairwiseMNISTsmallweights} for $M=100$, $M=400$ and $M=1200$. It is clear from the figure that the small parameter approximation provides pretty good estimates of pairwise interactions in all cases except for small values of $M$. In fact, as demonstrated by Figure~\ref{fig:pairwiseMNISTsmallweights2}, the typical strength of RBM weights measured by the root mean square value $\sqrt{\frac{1}{NM}\sum_{i,\mu}w_{i,\mu}^2}$, decreases with $M$ for all activation functions, which makes the approximation more and more accurate, as in fact confirmed by corresponding decreasing values of the normalised root mean square error $\sqrt{\frac{\sum_{i < j} (J_{ij}-\tilde{J}_{ij})^2}{\sum_{i < j} J_{ij}^2}}$, where $J$ denotes pairwise interactions obtained with Eq.\ \eqref{eqn:general_int_terms_factorized} and $\tilde{J}$ their approximation according to Eq.\ \eqref{eqn:small_w_approximation}. The reason why the magnitude of the inferred weights decreases with the size of the hidden layer is related to the fact that the learning algorithm must operate in a region where parameters can be easily tuned in order to efficiently learn data representation. In fact, small values of the weights help preventing the formation of large input currents, especially for large values of $M$, which might instead saturate network dynamics and stop training. 

\begin{figure}
\begin{center}
\includegraphics[width=.7\columnwidth]{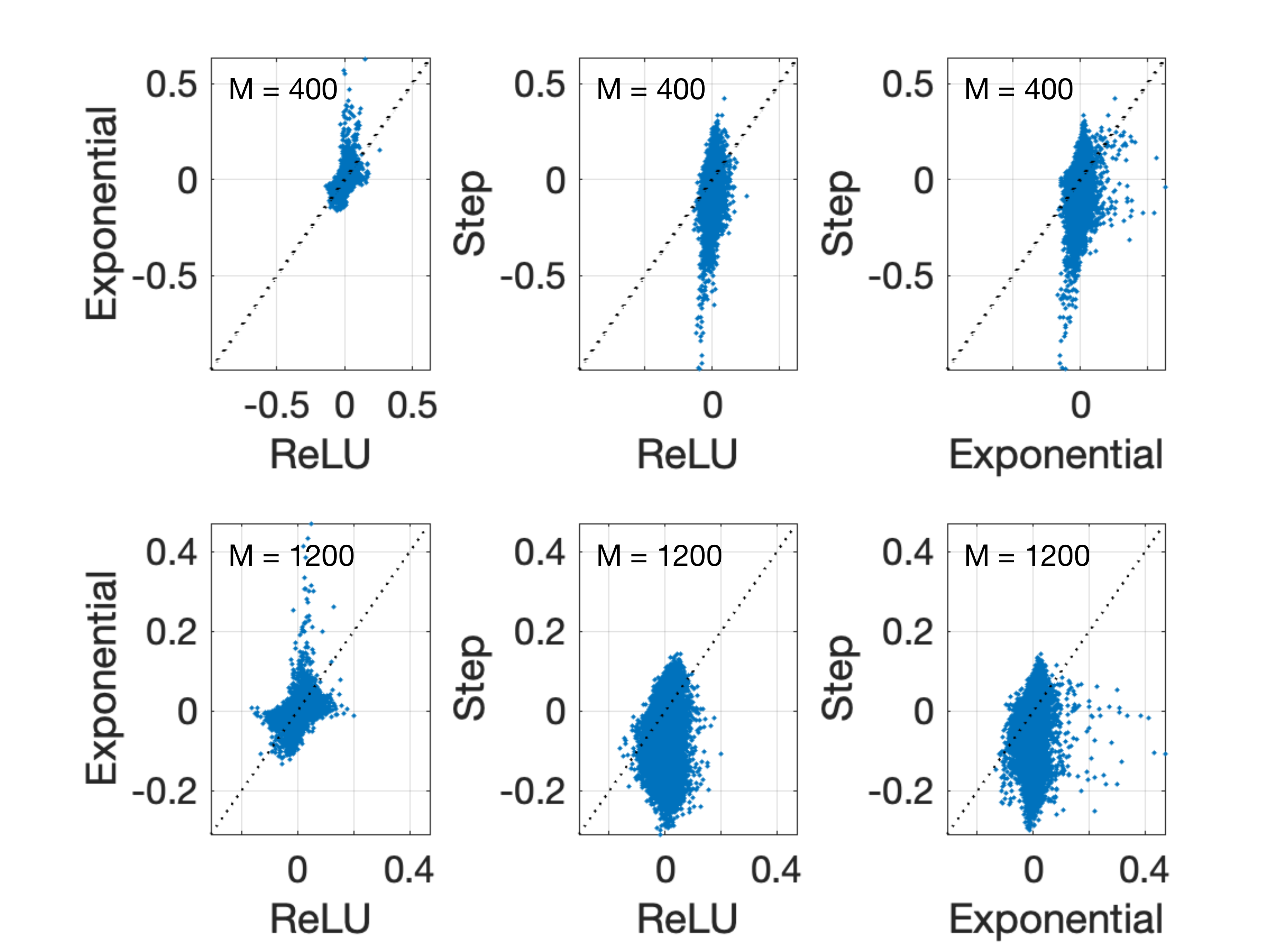}
\caption{Comparison across values of the three-body interaction terms as induced by RBMs with different non-linear activation functions for $M=400$ (top row) and $M=1200$ (bottom row).}
\label{fig:three_body}
\end{center}
\end{figure}

So far we have seen that, as far as the fields and pairwise interactions induced by the hidden layer on the visible ones are concerned, the different activation functions do lead to different parameters, but there are also clear scaling and linear relationships between them, at least when there is a large enough hidden layer. In fact, in this case, the small parameter approximation for the pairwise interactions holds with a good accuracy and, within this approximation, differences across activation functions in the induced parametrisation merely reduce to factors involving cumulants of the hidden node distributions (see Eq.\ \ref{eqn:ho_small_w_approximation}). However, at odds with the case of pairwise interactions, triplet interactions do not exhibit the same relationship among activation functions, as shown in Figure~\ref{fig:three_body} where we compare the triplet interactions induced by hidden layers across non-linear activation functions for $M=400$ and $M=1200$ (in the linear case, the three-body interactions terms are trivially zero). These differences observed for the triplet interactions might be partly ascribed to the fact that values of higher order interactions are expected to be relatively more noisy, as we will argue in the following.

\begin{figure}
\begin{center}
\includegraphics[width=.5\columnwidth]{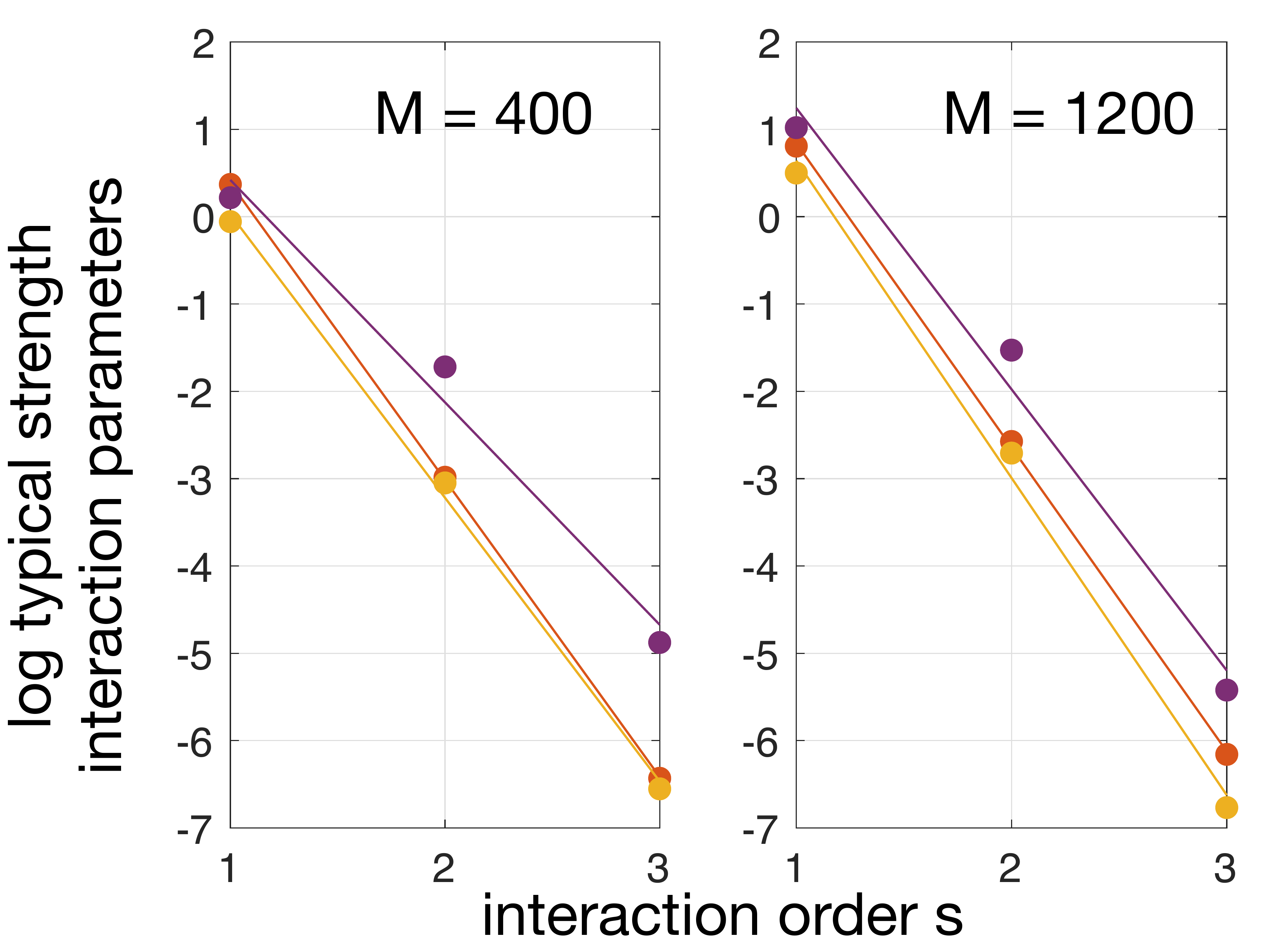}
\caption{The typical values of the interaction strength (the logarithm of their root mean square value) induced by RBMs with ReLU (red), Exponential (yellow) and Step (violet) activation functions plotted versus the interaction order (fields $s=1$, pairwise interactions $s=2$ and three-body interactions $s=3$) for $M = 400$ and $M=1200$. For each activation function in each subfigure a linear regression line is drawn.}\label{fig:figure11}
\end{center}
\end{figure}

The typical values of the interaction strength induced by RBMs with different non-linear activation functions are shown in Figure~\ref{fig:figure11} versus the interaction order $s$ for $M = 400$ and $M=1200$. In the figure, $s=1$ refers to the field parameters, $s=2$ to the pairwise interactions and $s=3$ to the three-body terms, whereas the interaction strength is estimated as the logarithm of the root mean square value of corresponding interaction parameters. For what concerns the field parameters, we are considering in the figure only the contribution coming from the RBM weight parameters, that is, we are not including the bias term (see Eqs.\ \ref{eqn:general_int_terms_factorized} and \ref{eqn:4int_terms_separable}) so that we can compare on the same footing parameters induced by the hidden layer across different order. Figure~\ref{fig:figure11} reveals that the magnitude of interactions parameters are typically smaller in ReLU and Exponential as compared to Step, a feature which is likely to be due to the fact that the latter can only output values in a small finite range at odds with the former. It also displays that fields and pairwise interactions grows with $M$, as already observed in previous figures. Most interestingly, the strength of interaction terms decreases with the interaction order following a trend that is consistent with an exponential decay (see regression lines in the figure; see also distributions of RBM parameters and interaction terms in section 2.1 of the Supplemental Material). This is, in fact, what one would expect assuming that parameters are small enough and considering that, within this assumption, the first contribution to an interaction term of order $s$ would be roughly $I^{(s)} \sim w^s$ with $w$ representing the typical weight strength, as we have indeed verified in Figure~\ref{fig:pairwiseMNISTsmallweights} and Figure~\ref{fig:pairwiseMNISTsmallweights2} for the pairwise terms. As a matter of fact, the slopes of the curves in Figure~\ref{fig:figure11} imply values of $w$ around $w\sim 0.035$ (except for the Step case with M = 400) which are roughly of the same order of the typical RBM weights, as illustrated in Figure~\ref{fig:pairwiseMNISTsmallweights2}. All together these observations point to the fact that, in the case of the MNIST dataset, the interaction models learnt by different RBMs would all present a hierarchical structure where lower order parameters are more relevant then higher order ones, a feature we have already observed in smaller RBMs in section \ref{sec:numerical_verification} (Figure~\ref{fig:RBMexample2}) which was caused by the small values of RBM weights and which indeed disappeared with increasing parameter strength (Figure~\ref{fig:RBMexample3}).

Besides getting smaller, values of higher order parameters are also expected to be more sensibly affected by the statistical error of RBM parameter estimation. In fact, assuming for simplicity that $I^{(s)} \sim w^s$ then the statistical error on the estimate of $w$, that is $\delta_w$, would determine an uncertainty on the induced interaction terms which grows exponentially with the interaction order, that is $\frac{\Delta I^{(s)}}{I^{(s)}} \sim (1+\frac{\delta_w}{w})^s - 1$. It is easy to verify that this result holds also when using Eq.\ \eqref{eqn:ho_small_w_approximation}, instead of the simple equation $I^{(s)} \sim w^s$, and within the assumption that the relative error estimate is approximatively the same across weight parameters and equal to $\delta_w/w$. A relative increase in the uncertainty of triplets with respect to lower order parameters is in fact observed in section 2.2 of the Supplemental Material where we report the comparison between interaction parameters retrieved from two halves of the dataset. As anticipated, this fact might also have contributed to the larger differences among triplet interactions found in Figure~\ref{fig:three_body}. 

Finally, the decrease of the typical value of interaction parameters with the interaction order, together with the increase of the relative uncertainty, sets a limit on the maximum order of interaction which make sense to consider, that is, an upper bound on the complexity of the interaction model that the RBM is representing. In fact, by requiring that $\frac{\Delta I^{(s)}}{I^{(s)}} \sim (1+\frac{\delta_w}{w})^s - 1 \ll 1$, we find that $s \ll \frac{\log 2}{\log(1+\frac{\delta_w}{w})} \sim 0.7 \frac{w}{\delta_w}$. For instance, if the error is $\frac{\delta_w}{w} \sim 10\%$ then the interaction orders that would make sense to consider are such that $s \lesssim 7$: higher order interactions would not be meaningful given that their estimate is not robust against statistical error.

In conclusion, in this section we have seen that RBMs trained on the MNIST dataset with a standard learning algorithm learn interaction models which are quite consistent across nonlinear activation functions and hidden node number. Furthermore these models present a hierarchy of interactions where lower order interactions are typically attached larger absolute values and their estimates are not much affected by statistical errors, while higher order interactions are. This is especially true when inferred RBM weights are small enough, a condition which is almost always satisfied in the simulations, except for small $M$. We will discuss this issue later in the Discussion. 

\section{Summary}
In this paper we studied the marginal distribution over the visible nodes of Restricted Boltzmann Machines for different hidden layer activation functions and number of hidden nodes. We did this by deriving an explicit expression for the marginal distribution over the observed variables as a model of interacting binary variables, reminiscent of spin models studied in statistical physics, and then studying the interaction terms of different orders that appear in this representation. These interactions, explicitly written down in Eq.\ \eqref{eqn:4int_terms_separable}, ends up to have a rather straightforward and intuitive structure as a function of the cumulant generating functional $K$ defined in Eq.\ \eqref{eq:Kdef} for any hidden node potential for which the probability $\rho$ defined in Eq.\ \eqref{eqn:density} is well defined. In order to derive the mapping for a given choice of the potential, one only needs to calculate the cumulant generating function which is known for many distributions and, can be easily evaluated for separable potentials. 

The mapping correctly predicts that when the Linear activation function is selected, the corresponding RBM can only describe pairwise models, at odds with the other investigated cases. In particular, we have demonstrated that the Linear activation function is indeed the only possible choice which induces a model with only a finite set of nonzero interactions given a generic RBM weight matrix. Moreover we have shown that the RBM with the Linear activation function provides the best low rank approximation to any pairwise model and, in particular, can express a pairwise model with a full rank connectivity matrix (Ising model) when the number of hidden nodes $M$ equals the number of visible minus one. The same efficiency is inherited by RBMs with ReLU \citep{nair2010rectified}, which can be approximately interpreted as a semi-Restricted Boltzmann Machine given the similarities of the higher order interaction terms with the standard RBM. In both ReLU and Exponential, trivial symmetries in the pairwise representation are broken through the presence of higher order terms which are modelled by the same parameters, promoting the interpretability of the emerging latent variable representation. Interestingly, in the small parameter limit, pairwise interactions will always be parametrised by a Hopfield-like term, regardless of the choice of the potential (Eq.\ \ref{eqn:small_w_approximation}). Finally, the Exponential case \citep{welling2005exponential,gehler2006rate} shows a surprisingly simple representation of higher order terms which seems to generalise the efficient representation employed for pairwise interactions. Analogously to the Hopfield case, such a representation is shared across activation functions in the small parameter limit (Eq.\ \ref{eqn:ho_small_w_approximation}).

By performing numerical tests, we have verified that the expressions we report in this paper describe correctly the interactions among the observed nodes in a number of simple RBM architectures. As an application, we have studied the interactions induced in RBMs trained on the MNIST data finding simple scaling relationships between pairwise interactions resulting from RBMs with different number of hidden nodes and activation functions. The same is true for the fields parameters whereas we did not observe any clear pattern for the triplet interactions. In almost all investigated cases, except for small hidden node layer, the RBM weight parameters are small enough to justify a small parameter approximation: we have checked the approximation for the pairwise interaction parameters, that is Eq.\ \eqref{eqn:small_w_approximation}, holds with a good accuracy and we have verified that the magnitude of the typical interactions decrease with the interaction order as Eq.\ \eqref{eqn:ho_small_w_approximation} would predict and as observed also in numerical experiments. Furthermore, by inspection of Eq.\ \eqref{eqn:ho_small_w_approximation}, we have also noticed that interaction models described by the RBMs present an increasing uncertainty in the estimates of the interaction parameters which grows with the interaction order and which implicitly sets an upper bound on the largest interaction order that the RBM can represent. 

\section{Discussion}
Our study is related to several previous works where a direct or indirect link can be traced between generalised RBMs and spin models.
The expansion of log-probability distribution of RBMs in terms of disjoint events reminds of the method of Network Polynomials in Bayesian networks where a number of probabilistic queries could be retrieved by evaluating partial derivatives of such a polynomial \citep{darwiche2003differential}. Furthermore, the mapping presented here can also be seen as an application to RBMs of the $\chi$-expansion proposed in \citet{martignon1995detecting} for detecting higher-order interactions in neural data. 

\citet{martens2010parallelizable} exploit the equivalence between pairwise Markov Random Fields (MRF) and RBMs with gaussian hidden units in order to introduce a parallelisable scheme for efficient sampling of hard-to-sample MRFs. On the same line, and very related to our work is the analysis reported in \citet{yoshioka2019transforming}. The authors in this work address the question that is the inverse of what we considered here, namely going from an interacting binary model distribution to an equivalent RBM in the case of binary hidden units. They report an algebraic transformation of generalized Ising models with many-spin interactions into Boltzmann machines, exploring the benefit of the mapping in Monte Carlo simulations. 

In a set of papers, \citet{barra2012equivalence,barra2018phase} studied the phase diagram of RBMs by varying the prior over weight parameters and the prior over node variables, i.e. the activation function, and focus on the conditions for the existence of a retrieval phase. From the point of view of statistical learning, the authors argue that the retrieval phase is associated with models that after training are capable of capturing few relevant features from heterogeneous data while avoiding overfitting and that the paramagnetic to spin glass transition conveys information on the optimal size of the training set required for making good inferences \citep{barra2017phase}; see also \citep{marullo2021boltzmann} for a recent review of some of these results as well as the relationship between dense associative networks and RBMs. The role of the non-linearity in the activation function in the expressive power of the RBMs has also been studied in \citep{tubiana2017emergence} and \citep{tubiana2019learning} via numerical experiments and a replica analysis. In this work, the conditions under which RBMs enter the so-called ``compositional phase'' where visible configurations are composed from combinations of a number of features encoded by simultaneously strongly activated hidden units was studied. RBMs working in such a regime can capture invariant features in the underlying distribution from vastly distinct data configurations while the sparse activation of the hidden layer favours their interpretation. Although all these studies address the same general question as the one studied here, namely the relationship between the representational power of RBMs and features of the activation function and connectivity employed, they differ from ours in that their analyses are performed on ensembles of RBMs analysed using statistical mechanics techniques in the thermodynamics limit. In other words, the analyses aim at understanding the typical case behaviour of an RBM within a given statistical ensemble. Our focus on the other hand has been to directly relate the marginal distribution of the observed nodes to that of the activation function for a single realisation of the system. It would, however be interesting to see if applying the statistical mechanics approaches used in previous studies such as \citet{barra2017phase,tubiana2017emergence}, combined with the mapping studies here as well as the existing literature on the phase diagram of the $p$-spin glass models (see \citet{barrat1997p} for a review) may allow identification of the phase diagram and, in particular, the most expressive regime of RBMs with arbitrary activation functions. In this case, the small coupling expansion in Eq.\ \eqref{eqn:ho_small_w_approximation} could be a useful approximation to simplify the analytical calculations.

The observations summarised in the previous section on RBMs trained on the MNIST suggest that with all the nonlinear activation functions, the interactive marginal distribution of the observed nodes is a low order interaction model with a hierarchy in the relative strength of interaction parameters that outperform simple pairwise models only when providing a large number of hidden nodes. Although the emergence of this hierarchy in the interactions might partly be due to the actual data structure, in general, this might also be the result of the learning algorithm that output parameters which are typically small in magnitude. In particular, parameter initialisation, small learning rates and regularisation favour the formation of this hierarchy by promoting small parameter values which, as already noticed, is a necessary condition when dealing with unbounded and strictly positive activation functions such as the ReLU and the Exponential ones in order to discourage the formation of big inputs which could cause divergences during training.
%

Beside providing insights on the way a given RBM represents probability distributions of binary variables, this work shows that RBMs can be potentially useful for graphical model selection in models of binary variables: by controlling the degree of connectivity of the hidden nodes and the type of the activation function, it is possible to use the mapping to limit the maximum order of interactions described by the RBM. In this way, one could verify, with the same RBM, the potential contribution of different order of interactions to the data representation. For instance, one could test on data, pairwise models against models employing also three-body interaction terms by comparing two RBMs of the same type but with hidden nodes connected to at most two visible nodes in the first case and three visible nodes in the second and so on. 

Finally, we find it an interesting avenue of research to see whether given some knowledge about the statistics of the datasets and tasks, proper activation functions could be chosen or even learnt so as to be useful for that dataset or tasks. In particular, given recent observations about the effect of activation function on the efficiency of simple learning rules e.g. in \cite{schonsberg2021efficiency}, it would be interesting to see if simple learning rules and algorithms can become more efficient for particular tasks and datasets provided that the hidden layer activation function is chosen not without regards to the data (as is largely done today), but chosen or even learned focusing on what a given activation function can and cannot represent. 

\section*{Acknowledgments}
This work has been supported by the Research Council of Norway (Centre for Neural Computation, grant number 223262; NORBRAIN1, grant number 197467), and the Kavli Foundation.
 
\medskip
\bibliographystyle{apa}
\bibliography{mybibAPS}{}

\appendix
\section{Activation functions and their cumulant generating functions}
\label{sec:Activationfunctions}
\subsection{Linear activation function}
an RBM with a quadratic separable potential 
\begin{equation}\label{eqn:linpot}
U_\mu(z_\mu) = \frac{z_\mu^2}{2} + c_\mu z_\mu \quad \forall \mu.
\end{equation}
is also known as a Bernoulli-Gaussian RBM. The conditional probability distribution $P(z_\mu|v)$ is Gaussian and centred around $\langle z_\mu \rangle = I_\mu - c_\mu$ with unit variance:
\begin{align}
P(z_\mu |v) = \frac{\exp\{-\frac{1}{2}[z_\mu-(I_\mu - c_\mu)]^2\}}{\sqrt{2\pi}}.
\end{align}
Since for a gaussian distribution the mode $\overline{z}_\mu $ is equal to the mean, it follows that, in this case, the activation function is a linear function of the total input to each hidden node. Moreover, by comparing Eq.\ \eqref{eqn:C_conditional_probability_separable} and Eq.\ \eqref{eqn:density}, it is clear that the density $\rho_\mu (z_\mu)$ is equal to $P(z_\mu|v)$ when $\bv\tr W_\mu=I_\mu = 0$, implying that $\rho_\mu (z_\mu)$ is also Gaussian and centred around $-c_\mu$. Using Eq.\ \eqref{eqn:linpot} in Eq.\ \eqref{eqn:integral}, the cumulant generating functions; for the Linear activation function can thus be written as
\begin{equation}
K_\mu(q_\mu) = \frac{q_\mu^2}{2} - q_\mu c_\mu.
\end{equation}
Cumulants are by definition derivatives of the cumulant generating function evaluated at zero. In this case, since the distribution $\rho_\mu (z_\mu)$ is Gaussian, only the first two cumulants are different from zero, that is $\kappa_\mu^{(1)} = -c_\mu$ and $\kappa_\mu^{(2)} = 1$.

\subsection{ReLU activation function}
A ReLU activation function is obtained assuming a potential that is quadratic only for positive values of its argument
\begin{equation}
U_\mu(z_\mu) = 
\begin{cases}
\frac{z_\mu^2}{2} + c_\mu z_\mu \quad & z_\mu \geq 0 \\
+\infty  & z_\mu < 0 .
\end{cases}
\end{equation}
In fact, in this case, the mode of the conditional distribution $P(z_\mu|v)$, which can be written as
\begin{equation}
P(z_\mu|v) =
\begin{cases}
\frac{\exp\{-\frac{1}{2}[z_\mu-(I_\mu - c_\mu)]^2\}}{\int_0^{+\infty} dz_\mu \exp\{-\frac{1}{2}[z_\mu-(I_\mu - c_\mu)]^2)\}} \qquad &\forall z_\mu \geq 0 \\
0 \qquad &\forall z_\mu < 0
\end{cases}
\end{equation}
is equal to $\overline{z}_\mu = \max(0,I_\mu- c_\mu)$. The mean of the distribution here is different from the mode and resembles a softplus activation function $\langle z_\mu \rangle = I_\mu -c_\mu +\sqrt{2/\pi}\frac{\exp[-(I_\mu - c_\mu)^2/2]}{1+\erf{[(I_\mu- c_\mu)/\sqrt{2}]}}$. As before, $\rho_\mu (z_\mu)$ is equal to $P(z_\mu|v)$ when $I_\mu = 0$ and by performing the integral in Eq.\ \eqref{eqn:integral}, one can derive the cumulant generating function
\begin{equation}\label{eqn:cum_relu}
K_\mu(q_\mu) = \frac{q_\mu^2}{2} - q_\mu c_\mu + \log\frac{1+\erf[(q_\mu-c_\mu)/\sqrt{2}]}{1-\erf(c_\mu/\sqrt{2})},
\end{equation}
where $\erf(x)$ is the error function. In the numerical experiments in section \ref{sec:numerical_experiments}, we compared the pairwise terms obtained by applying the mapping of Eq.\ \eqref{eqn:general_int_terms_factorized} against the same parameters as evaluated through the simpler expression in Eq.\ \eqref{eqn:small_w_approximation} which is exact only in the limit of small parameters. In order to determine the values of the interaction parameters through Eq.\ \eqref{eqn:small_w_approximation}, one needs to calculate the second cumulant of the distribution $\rho_\mu (z_\mu)$. In the case of ReLU, by differentiating twice Eq.\ \eqref{eqn:cum_relu} and evaluating the result in zero, one gets
\begin{equation}
\kappa^{(2)}_\mu = 1 + \sqrt{\frac{2}{\pi}}\frac{\exp(-c^2/2)}{1-\erf(c/\sqrt{2})}\left[c - \sqrt{\frac{2}{\pi}}\frac{\exp(-c^2/2)}{1-\erf(c/\sqrt{2})}\right].
\end{equation}

\subsection{Step activation function}
The standard RBM, or Bernoulli-Bernoulli RBM, is retrieved when employing the following potential
\begin{equation}
U_\mu(z_\mu) = 
\begin{cases}
c_\mu z_\mu \quad & z_\mu \in \{0,1\} \\
+\infty  & \mbox{otherwise}.
\end{cases}
\end{equation}
In fact, in this case, the hidden nodes are binary variables and the conditional probability distribution $P(z_\mu|v)$ is a sigmoid function
\begin{equation}
P(z_\mu|v) =
\begin{cases}
\frac{\exp[z_\mu(I_\mu - c_\mu)]}{1 + \exp(I_\mu - c_\mu)}\qquad & z_\mu \in \{0,1\} \\
0 \qquad & \mbox{otherwise}.
\end{cases}
\end{equation}
The mode of the distribution is a step function of the total input minus the threshold, $\overline{z}_\mu = \Theta(I_\mu- c_\mu)$ where $\Theta(x)$ is the Heaviside function, whereas the mean of the distribution is a sigmoid function $\langle z_\mu \rangle = \frac{\exp(I_\mu - c_\mu)}{1 + \exp(I_\mu - c_\mu)}$. It follows from Eq.\ \ref{eqn:integral} that the cumulant generating function is equal to\begin{equation}
K_\mu(q_\mu) = \log\frac{1+\exp(q_\mu-c_\mu)}{1+\exp(-c_\mu)}.
\end{equation}
Finally the second cumulant of the distribution is given by the expression
\begin{equation}
\kappa^{(2)}_\mu = \frac{\exp(-c_\mu)}{(1+\exp(-c_\mu))^2}.
\end{equation}

\subsection{Exponential activation function}
An Exponential activation function is induced by the following potential
\begin{equation}
U_\mu(z_\mu) = 
\begin{cases}
c_\mu z_\mu + \log z_\mu! \quad & \forall z_\mu \in \mathbb{N} \\
+\infty & \mbox{otherwise}.
\end{cases}
\end{equation}
Consequently, the conditional distribution is a Poisson distribution
\begin{equation}\label{eqn:conditional_distribution_poisson}
P(z_\mu|v) =
\begin{cases}
e^{-\lambda}\frac{{\lambda}^{z_\mu}}{z_\mu !} \qquad & z_\mu \in\mathbb{N} \\ 
0 \qquad &\mbox{otherwise}.
\end{cases}
\end{equation}
with a parameter $\lambda_\mu = \exp(I_\mu - c_\mu)$. The mean of the distribution is $\lambda_\mu$, that is $\langle z_\mu\rangle = \exp(I_\mu - c_\mu)$ and the mode, $\overline{z}_\mu = \floor{(\exp(I_\mu - c_\mu))}$, where the floor function approximates its argument to the closest smaller integer. Since $\rho_\mu(z_\mu)$ is equal to $P(z_\mu|v)$ when $I_\mu = 0$, it follows that $\rho_\mu(z_\mu)$ is also a Poisson distribution with parameter $\tilde{\lambda}_\mu = \exp(-c_\mu)$. The cumulant generating function for such a distribution is given by \begin{equation}
K_\mu(q_\mu) = \exp(-c_\mu) [\exp(q_\mu)-1]
\end{equation}
and it is easy to verify by differentiating the above expression that all cumulants are equal to $\tilde{\lambda}$.

\end{document}